\pdfoutput=1
\documentclass{article}

\usepackage[preprint, nonatbib]{neurips_2023}

\usepackage[utf8]{inputenc} % allow utf-8 input
\usepackage[T1]{fontenc}    % use 8-bit T1 fonts
\usepackage{hyperref} 
\hypersetup{breaklinks,colorlinks}
\usepackage{url}            % simple URL typesetting
\usepackage{booktabs}       % professional-quality tables
\usepackage{amsfonts}       % blackboard math symbols
\usepackage{nicefrac}       % compact symbols for 1/2, etc.
\usepackage{graphicx}
\usepackage{microtype}      % microtypography
\usepackage{xcolor}         % colors
\usepackage{amsmath}
\usepackage{wrapfig}
\usepackage{multirow}
\newcommand{\ut}[1]{{\color{black} #1}}

\newcommand{\yj}[1]{{\color{black} #1}}

\DeclareMathOperator{\sign}{sign}

\title{What Knowledge Gets Distilled in Knowledge Distillation?}

\author{Utkarsh Ojha\thanks{Equal contribution} \hspace{20pt} Yuheng Li\footnotemark[1] \hspace{20pt} Anirudh Sundara Rajan\footnotemark[1] \\\\ \textbf{Yingyu Liang} \hspace{20pt} \textbf{Yong Jae Lee} \\\\
 University of Wisconsin-Madison\\
}

\begin{document}

\maketitle

\begin{abstract}
%\as{Should the abstract talk about our discovery of "What Knowledge" gets transferred?}
Knowledge distillation aims to transfer useful information from a teacher network to a student network, with the primary goal of improving the student's performance for the task at hand. Over the years, there has a been a deluge of novel techniques and use cases of knowledge distillation. Yet, despite the various improvements, there seems to be a glaring gap in the community's fundamental understanding of the process. Specifically, what is the knowledge that gets distilled in knowledge distillation?  In other words, in what ways does the student become similar to the teacher? Does it start to localize objects in the same way? Does it get fooled by the same adversarial samples? Does its data invariance properties become similar? Our work presents a comprehensive study to try to answer these questions. 
We show that existing methods can indeed indirectly distill these properties beyond improving task performance. We further study why knowledge distillation might work this way, and show that our findings have practical implications as well.
\end{abstract}

\vspace{-5pt}
\section{Introduction}
\vspace{-5pt}

%(typically large) 
%(typically small) 
Knowledge distillation, first introduced in~\cite{bucila-kdd2006,hinton-neurips2014}, is a procedure of training neural networks in which the `knowledge' of a teacher is transferred to a student. The thesis is that such a transfer (i) is possible and (ii) can help the student learn additional useful representations. The seminal work by \cite{hinton-neurips2014} demonstrated its effectiveness by making the student imitate the teacher's class probability outputs for an image. This ushered in an era of knowledge distillation algorithms, including those that try to mimic intermediate features of the teacher \cite{romero-iclr2015, at-iclr2017, huang-arxiv2017}, or preserve the relationship between samples as modeled by the teacher \cite{rkd-cvpr2019, fsp-cvpr2017}, among others. 
% as an alternative
%of this idea

While thousands of papers have been published on different techniques and ways of using knowledge distillation, there appears to be a gap in our fundamental understanding of it.  Yes, it is well-known that the student's performance on the task at hand can be improved with the help of a teacher.  But what exactly is the so-called \emph{knowledge} that gets distilled during the knowledge distillation process?  For example, does distillation make the student look at similar regions as the teacher when classifying images? If one crafts an adversarial image to fool the teacher, is the student also more prone to getting fooled by it? If the teacher is invariant to a certain change in data, is that invariance also transferred to the student? 
%Does the teacher's behavior on out-of-distribution data get transferred to the student?  
%If the teacher is shape-biased, does this property get transferred to the student, who otherwise would have been texture-biased? 
Such questions have not been thoroughly answered in the existing literature.
%, and thus, our goal is to shed some light on the `dark knowledge' \cite{dark_knowledge} that gets distilled in knowledge distillation.
%If the student is trained to mimic the teacher's properties on one domain, will that also make the student and teacher behave similarly on some other domain? 

%There has remained, however, a common theme to almost all the work done so far; the goal is to improve the student's performance on a task of interest, with the help of a teacher. 
%When looking from the point of view of the teacher network, knowledge distillation can be seen as model compression; compressing the teacher into a student. From the student's perspective, knowledge distillation offers a source of additional information for training a network (e.g., soft probabilities from the teacher). Both ways of thinking, ultimately, involve the same training procedure.
This has become particularly relevant because there have been studies which present some surprising findings about the distillation process. \cite{cho-iccv2019} showed that performing knowledge distillation with a bigger teacher does not necessarily improve the student's performance over that with a smaller teacher, and thus raised questions about the effectiveness of the distillation procedure in such cases. \cite{stanton-neurips2021} showed that the agreement between the teacher and distilled student's predictions on test images is not that different to the agreement of those between the teacher and an independently trained student, raising further doubts about how knowledge distillation works, if it works at all.
In this work, we present a comprehensive study tackling the above questions.  We analyze three popular knowledge distillation methods \cite{hinton-neurips2014, romero-iclr2015, tian-iclr2020}. Many of our findings are quite surprising. For example, by simply mimicking the teacher's output using the method of \cite{hinton-neurips2014}, the student can inherit many implicit properties of the teacher. It can gain the adversarial vulnerability that the teacher has. If the teacher is invariant to color, the student also improves its invariance to color. To understand why these properties get transferred without an explicit objective to do so, we study the distillation process through a geometric lens, where we think about the features from a teacher as relative positions of an instance (i.e., distances) from its decision boundary. Mimicking those features, we posit, can therefore help the student inherit the decision boundary and (consequently) the implicit properties of the teacher. We show that these findings have practical implications; e.g., an otherwise \emph{fair} student can inherit biases from an unfair teacher. Hence, by shedding some light on the `dark knowledge' \cite{dark_knowledge},
our goal is to dissect the distillation process better.

\vspace{-5pt}
\section{Related work}\label{sec:ref}
\vspace{-5pt}

Model compression \cite{bucila-kdd2006} first introduced the idea of knowledge distillation by compressing an ensemble of models into a smaller network. \cite{hinton-neurips2014} took the concept forward for modern deep learning by training the student to mimic the teacher's output probabilities. Some works train the student to be similar to the teacher in the intermediate feature spaces \cite{romero-iclr2015, at-iclr2017}.  Others train the student to mimic the relationship between samples produced by the teacher \cite{rkd-cvpr2019, tung-iccv2019, peng-iccv2019}, so that if two samples are close/far in the teacher's representation, they remain close/far in the student's representation.  Contrastive learning has recently been shown to be an effective distillation objective in \cite{tian-iclr2020}. More recently, \cite{beyer-arxiv2021} present practical tips for performing knowledge distillation; e.g., providing the same view of the input to both the teacher and the student, and training the student long enough through distillation.
Finally, the benefits of knowledge distillation have been observed even if the teacher and the student have the same architecture \cite{born-again, dml}. For a more thorough survey of knowledge distillation, see \cite{gou-arxiv2021}. In this work, we choose to study three state-of-the-art methods, each representative of the output-based, feature-based, and contrastive-based families of distillation approaches.  

There have been a few papers that present some surprising results. \cite{cho-iccv2019} shows that a smaller \& less accurate teacher is often times better than a bigger \& more accurate teacher in increasing the distilled student's performance. More recent work shows that the agreement between the predictions of a teacher and student is not necessarily much higher than that between the teacher and an independent student \cite{stanton-neurips2021}. There has been work done which tries to explain why distillation \emph{improves student's performance} by trying linking it to the regularizing effect of soft labels \cite{kd_soft_reg, self_kd_soft} or what the key ingredients are which help in student's optimization process \cite{understand_kd, menon-icml}. What we seek to understand in this work is different: we study different ways (beyond performance improvement) in which a student becomes similar to the teacher by inheriting its implicit properties.
%However, even if the teacher and student do not become similar along one axis, they could still become more similar along different ones. Our study thus investigates the various other ways in which knowledge can get distilled into a student.
%Although no comprehensive study exists in trying to understand knowledge distillation, there have been a few papers that present some surprising results.  \cite{cho-iccv2019} challenge the assumption that a teacher with better test accuracy, which is typically a bigger network, would have more knowledge to pass on to the student.  Instead, it shows that distillation with a bigger teacher does not necessarily increase a student's accuracy compared to having a relatively smaller teacher.

\vspace{-5pt}
\section{Distillation methods studied}\label{sec:methods}
\vspace{-5pt}
%Representative distillation methods studied

\begin{figure*}[t]
    \centering
    \includegraphics[width=.98\textwidth]{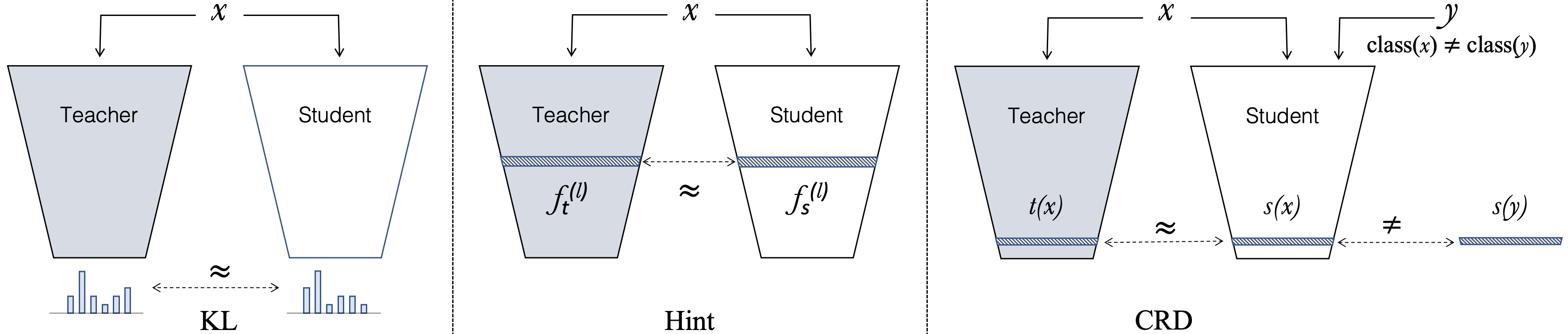}
    \vspace{-3pt}
    \caption{Methods used in this work. (i) $KL$ \cite{hinton-neurips2014}: mimicking of class probabilities. (ii) $Hint$ \cite{romero-iclr2015}: mimicking of features at an intermediate layer. (iii) $CRD$ \cite{tian-iclr2020}: features from the student and teacher for the same image constitute a positive pair, and those from different classes make up a negative pair.}
    \label{fig:methods}
    \vspace{-4pt}
\end{figure*}

%in this study 
%and popular 
To ensure that our findings are general and cover a range of distillation techniques, we select standard methods representative of three families of distillation techniques: output-based~\cite{hinton-neurips2014}, feature-based~\cite{romero-iclr2015}, and contrastive-based~\cite{tian-iclr2020}. The objectives of these methods, described below, are combined with the cross entropy loss $\mathcal{L}_{\mathrm{CLS}}(\mathbf{z_s}, \mathbf{y}) := -\sum_{j=1}^c y_j \log\sigma_j(\mathbf{z_s})$, where $\mathbf{y}$ is the ground-truth one-hot label vector, $\mathbf{z_s}$ is the student's logit output, $\sigma_j(\mathbf{z}) = \exp(z_j) / \sum_i \exp(z_i)$ is the softmax function, and $c$ is the number of classes.  
% (before softmax)

%The objectives of these methods, described in detail shortly, are used in addition to the standard cross entropy loss $\mathcal{L}_{\mathrm{CLS}}(\vec z_s, \vec y) := -\sum_{j=1}^c y_j \log\sigma_j(\vec z_s)$, where $\vec y$.  is the ground-truth binary label, $\vec z_s$ is the logit output of the student (before softmax), $\sigma$ is the softmax function, and $c$ is the number of classes.   

(1) \textbf{KL:} \cite{hinton-neurips2014} proposed to use the soft labels produced by the teacher as an additional target for the student to match, apart from the (hard) ground-truth labels. This is done by minimizing the KL-divergence between the predictive distributions of the student and the teacher: 
\vspace{-1pt}
\begin{align}
    \mathcal{L}_{\mathrm{KL}}(\mathbf{z_s}, \mathbf{z_t}) :=
    - \tau^2 \sum_{j=1}^c \sigma_j\left(\frac{ \mathbf{z_t}}{\tau}\right) \log \sigma_j\left(\frac{ \mathbf{z_s}}{\tau}\right), %\hspace{6mm}   \sigma_i\left(\frac{ \vec z}{\tau}\right) = \frac{\exp(z_i/T)}{\sum_j \exp(z_j/T)}
    \label{eq:kl}
\end{align}
where $\mathbf{z_t}$ is the logit output of the teacher, and $\tau$ is a scaling temperature. The overall loss function is $\gamma\mathcal{L}_{\mathrm{CLS}} + \alpha\mathcal{L}_{\mathrm{KL}}$, where $\gamma$ and $\alpha$ are balancing parameters. We refer to this method as $KL$.
%scaled by an appropriate

%Since there will typically be a discrepancy between the feature dimensions of the teacher and those of the student, FitNets
%. The objective function is 
%between the two features
(2) \textbf{Hint:} FitNets \cite{romero-iclr2015} makes the student's intermediate features ($\mathbf{f_s}$) mimic those of the teacher's ($\mathbf{f_t}$) for an image $x$, at some layer $l$. It first maps the student's features (with additional parameters $r$) to match the dimensions of the teacher's features, and then minimizes their mean-squared error:
%\vspace{-2pt}
\begin{align}
    \mathcal{L}_{Hint}(\mathbf{f_s^{(l)}},\mathbf{f_t^{(l)}}) = \frac{1}{2} || \mathbf{f_t^{(l)}} - r(\mathbf{f_s^{(l)}}) ||^2 \label{eq:hint}
\end{align}
The overall loss is $\gamma\mathcal{L}_{\mathrm{CLS}} + \beta\mathcal{L}_{\mathrm{Hint}}$, where $\gamma$ and $\beta$ are balancing parameters. \cite{romero-iclr2015} termed the teacher's intermediate representation as $Hint$, and we adopt this name. 
%\ut{Unless otherwise specified, the layer $l$ is chosen to be somewhere in the middle for both networks. The original paper uses $Hint$ with $KL$, but in this work we are primarily interested in the role of individual objective functions.}

%are used to appropriately weigh the classification and distillation losses.  

(3) \textbf{CRD:} Contrastive representation distillation \cite{tian-iclr2020} proposed the following. Let $s(x)$ and $t(x)$ be the student's and teacher's penultimate feature representation for an image $x$. If $x$ and $y$ are from different categories, then $s(x)$ and $t(x)$ should be similar (positive pair), and $s(x)$ and $t(y)$ should be dissimilar (negative pair). A key for better performance is drawing a large number of negative samples $N$ for each image, which is done using a contantly updated memory bank.
%is the number of negative samples $N$ used for each image. \cite{tian-iclr2020} use a memory bank of all the training data, which is constantly updated, to draw a large number of negative samples on the fly: 
%\vspace{-1pt}
\begin{align}
    \mathcal{L}_{\mathrm{CRD}} = -\log h(s(x), t(x)) - \sum_{j=1}^N \log(1-h(s(x), t(y_j))) %\hspace{4mm} h(a, b) = \frac{e^{a \cdot b/\tau}}{e^{a \cdot b/\tau} + \frac{N}{M}} 
    \label{eq:crd}
\end{align}
where $h(a, b) = (e^{a \cdot b/\tau})/(e^{a \cdot b/\tau} + \frac{N}{M})$, $M$ is the size of the training data, $\tau$ is a scaling temperature, and $\cdot$ is the dot product. We use $CRD$ to refer to this method. 
%\yj{Do we use the cross entropy loss with this one?  If not, we need to change the description about it in the first paragraph of this section.}
All other implementation details (e.g., temperature for \emph{KL}, layer index for \emph{Hint}) can be found in appendix.% As one can see, these three algorithms are diverse enough, where students mimic different aspects of a teacher. 

%We chose these three methods because they are among the more popular distillation algorithms ($KL$, $Hint$), with $CRD$ being one of the recent and more effective method. 
%Also, each of them are different among themselves: $KL$ was proposed to handle features at the end of a network (class probabilities), $Hint$ proposed to make use of intermediate features, and $CRD$ focused on  

\vspace{-5pt}
\section{Experiments}\label{sec:exp}
\vspace{-5pt}

%\yh{Overall, I think it is better to includes the following two more studies (see the two scenarios in the intro): a model trained with softlabel (not from teacher), and it is not more similar to the teacher; a model distilled from one teacher, but top1 acc is not improved, but still similar to teacher in some ways. Maybe we can figure out that in which studies we can include these two cases.}
% use the image classification task as our case study, and
%In our analyses,
We now discuss our experimental design. 
%We  use ImageNet \cite{imagenet}, MNIST \cite{lecun-mnist1998}, as well as images from artistic domains \cite{geirhos2021partial}. 
%Unless otherwise mentioned, for all the analyses, the distillation procedure is performed using the training dataset, and evaluation is performed on the corresponding validation dataset. 
To reach conclusions that are generalizable across different architectures and datasets, and robust across independent runs, we experiment with a variety of teacher-student architectures, and tune the hyperparameters so that the distillation objective improves the test performance of the student compared to independent training.  For each setting, we average the results over two independent runs. 
%each time, the distilled student and independent student (without distillation) are trained using the same random seed. 
We report the top-1 accuracy of all the models in the appendix. The notation $\texttt{Net}_1 \rightarrow \texttt{Net}_2$ indicates distilling the knowledge of $\texttt{Net}_1$ (teacher) into $\texttt{Net}_2$ (student). 
%The network architectures for the teacher and the student are different for different analysis, and the reason for those choices are given in those respective sections. 

%As mentioned before, we are more interested in studying knowledge transfer between the teacher and the student in ways other than top-1 accuracy. Hence, the main text doesn't discuss the top-1 accuracy obtained either by the teacher or the student. For reference, those results can be found in the supplementary. In general, we do try to find the best configuration of hyper-parameters so that the resulting distilled student improves in accuracy over the independent student. 

\vspace{-2pt}
\subsection{Does localization knowledge get distilled?}
\vspace{-2pt}

\begin{figure*}[t]
    \centering
    \includegraphics[width=1\textwidth]{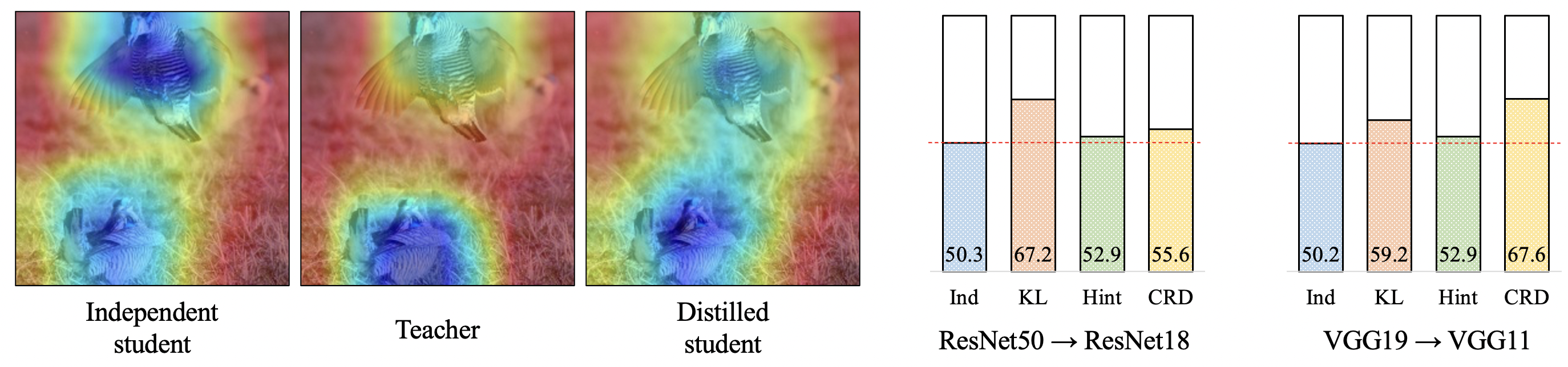}
    \vspace{-13pt}
    \caption{\textbf{Left:} An example of how the distilled student can focus on similar regions as the teacher while classifying an image. \textbf{Right:} \% where teacher's CAM is more similar to the distilled student's CAM than to the independent student's CAM. The red line indicates chance performance (50\%).}%color indicates the corresponding student's score. 
    \label{fig:cam}
    \vspace{-5pt}
\end{figure*}
%where 50\% (chance performance) lies
%\yh{Overall, I think we should include one transformer one in this study}

We start by studying whether the localization properties of a teacher transfers to a student through knowledge distillation.  For example, suppose that the teacher classifies an image as a cat by focusing on its face, and an independent student classifies it as a cat by focusing on its fur. After knowledge distillation, will the student focus \emph{more} on the face when classifying the image as a cat?  
%Different neural networks, especially of different architectures, will look at different regions of an image to classify it as one of the classes. Hence, the independent student and the teacher might have different class activation maps (CAMs). But does distillation push the map of the student to be closer to the teacher's? 

%use ImageNet for this experiment.
% (teacher or student)
\textbf{Experimental setup:} We train three models for ImageNet classification: a teacher, a distilled and an independent student. For each network, we obtain the class activation map (CAM) of the ground-truth class for each of random 5000 test images, using Grad-CAM \cite{grad-cam}, which visualizes the pixels that a classifier focuses on for a given class. We then compute how often (in terms of \% of images) the teacher's CAM is more similar (using cosine similarity) to the distilled student's CAM than to the independent student's CAM. A score of $>$50\% means that, on average, the distilled student's CAMs become more similar to those of the teacher than without distillation. As a sanity check, we also compute the same metric between the teacher's CAM and the CAMs of two independent students trained with different seeds (\emph{Ind}), which should be equally distant from the teacher (score of $\sim$50\%).
%$Ind$ computes the same frequency metric between the teacher's CAM and the CAMs of two independent students (both without distillation) trained with two random seeds. This is a sanity check baseline to show that these two models should be equally distant from the teacher, hence achieving a score of $\sim$50\%.  
%We compute this similarity through cosine similarity between the two activation maps. 
%Then, for each distillation method, w
%compares two random seeds of the independently trained student; 

\textbf{Results:} Fig.~\ref{fig:cam} (right) shows the results for two configurations of teacher-student architectures: (i) ResNet50 $\rightarrow$ ResNet18 and (ii) VGG19 $\rightarrow$ VGG11. For \emph{KL} and \emph{CRD}, we observe a significant increase in similarity compared to random chance (as achieved by the \emph{Ind} baseline).  \emph{Hint} also shows a consistent increase, although the magnitude is not as large.  
%We see that the activation maps of the ResNet-18 (student) produced through $KD$ distillation is twice as likely to be more similar to the teacher's, than an independent student's activation map. 

\textbf{Discussion:} This experiment shows that having access to the teacher's class-probabilities, i.e. confidence for the ground-truth class, can give information on where the teacher is focusing on while making the classification decision. This corroborates, to some degree, the result obtained in the Grad-CAM paper \cite{grad-cam}, which showed that if the network is very confident of the presence of an object in an image, it focuses on a particular region (Fig.~1(c) in \cite{grad-cam}), and when it is much less confident, it focuses on some other region (Fig.~7(d) in \cite{grad-cam}). 
%Not all distillation methods are created equal, however, and some have a lesser ability to make the activation map more similar to the teacher (e.g. $Hint$). 
Fig.~\ref{fig:cam} (left) shows a sample test image and the corresponding CAMs produced by the independent student (left), teacher (middle), and distilled student (right) for the ground-truth class. The distilled student looks at similar regions as the teacher, and moves away from the regions it was looking at initially (independent student). 
%Importantly, we are not interested in the correctness of any network's CAM; rather, we are using CAM as a tool to study whether a distilled student's CAMs become similar to those of the teacher's.  And our analysis indicates this to be the case for all three distillation techniques, albeit with varying degrees.  
So, regardless of the correctness of any network's CAM, our analysis shows that a distilled student's CAM does become similar to those of the teacher for all three distillation techniques, albeit with varying degrees.

%we do not wish to convey that the teacher's activation maps are better (though they might be for the task at hand), but rather to study whether the student's maps become similar to its, even if it is not the best map.
% should we talk about why we should study this, considering the reasons mentioned in the original KD paper about the teacher probably knowing more about the soft-classes in the image which the student might not know about, and hence we should test whether that hypothesis is actually happening (DISCUSS)

\vspace{-2pt}
\subsection{Does adversarial vulnerability get distilled?}\label{sec:adv}
\vspace{2pt}
%\as{Adversarially Robust Networks has technically done this before/}

\begin{figure*}[t]
    \centering
    \includegraphics[width=1\textwidth]{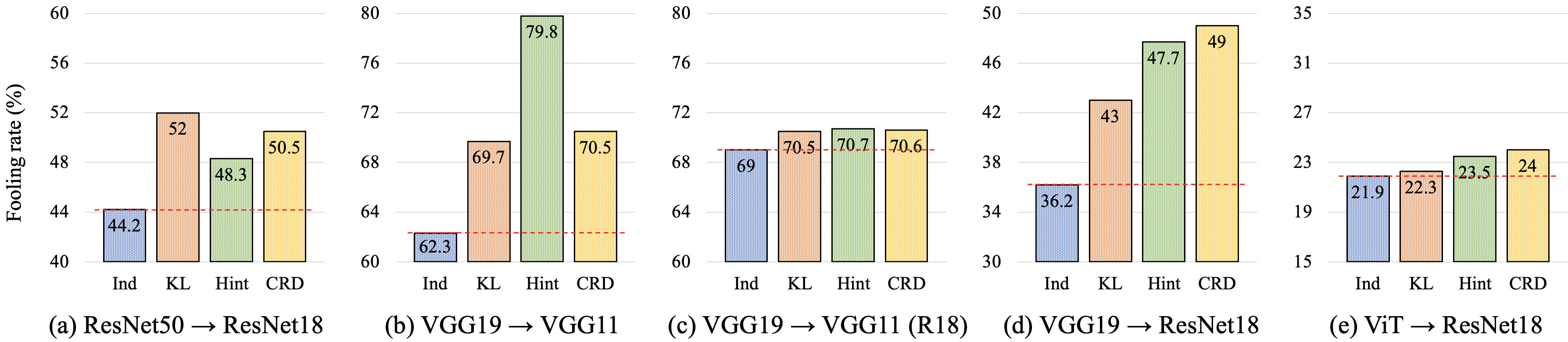}
    \vspace{-15pt}
    \caption{The images which fool the teacher fool the distilled student more than the independent student in \textbf{(a)}, \textbf{(b)} and \textbf{(d)}, but not in \textbf{(e)}. If the adversarial attack is generated using a foreign network (ResNet18, \textbf{(c)}), it fails to convincingly fool the distilled student more than the independent one. 
    %\textbf{(a, b} and \textbf{d)} 
    %\yh{I have a general concern for our plotting, in many cases, we actually want to make difference more obvious by only plot in certain scales, such as (a) (b). But this will make all difference more obvious than before. For example, in (d), at the first glimpse it seems hint improves a lot, but actually it is not true, only if we check numbers. So maybe for later case, we do not plot but rather just report numbers?} 
    }
    \label{fig:adv}
    \vspace{-5pt}
\end{figure*}

Next, we study the transfer of a different kind of property. If we design an adversarial image to fool the teacher, then will that same image fool the distilled student more than the independent student?
% Two different neural networks (teacher and independent student) will have their own separate ways of either correctly or incorrectly classifying an image. Now, in the previous section, we saw that the interpretability of the distilled students can become similar to that of the teacher when classifying correctly. Could this mean that the student \emph{misclassifies} in similar ways as the teacher as well? 
%\yh{I think for the previous section, we study how do networks make predication based on the regions they focus on, but this does not mean what is currently written, which read sounds like we do the following experiment: only for images correctly classified by all teacher, ind and distilled, sim(teacher, distilled) > sim(teacher, ind). Actually, I think maybe we can do the following study, for those images misclassfied by ind, but correctly classified by distilled and tea, to compare their grad-cam, to see if the difference it even bigger.}

\textbf{Experimental setup:} We train a teacher, a distilled student, and an independent student for ImageNet classification. Given 5000 random test images, we convert each image $I$ into an adversarial image $I^{adv}$ using iterative FGSM \cite{fgsm, iter-fgsm} (see appendix for details), whose goal is to fool the teacher, so that teacher's prediction for $I^{adv}$ changes from its original prediction for $I$. Fooling rate is then defined as the fraction of adversarial images which succeed at this task. In our experiments, we use only this fraction of adversarial images which fool the teacher, and apply them to different students.  

\textbf{Results:} We evaluate four configurations: (i) ResNet50 $\rightarrow$ ResNet18, (ii) VGG19 $\rightarrow$ VGG11, (iii) VGG19 $\rightarrow$ ResNet18, (iv) ViT (ViT-b-32) \cite{vit}  $\rightarrow$ ResNet18. 
%The attack is successful in $\sim$85\% of the cases for all the teachers. 
The fooling rate is $\sim$85\% for all the teachers.
Fig.~\ref{fig:adv} shows the fooling rate (y-axis) when applying these successful adversarial images to different types of students (x-axis).
We see that for ResNet50 $\rightarrow$ ResNet18, the ability to fool the independent student drops to 44.2\%, which is expected since the adversarial images aren't designed for that student.  In distilled students, we see an increase in the fooling rate relative to the independent one across all distillation methods (48\%-52\%). The trend holds for VGG19 $\rightarrow$ VGG11 and VGG19 $\rightarrow$ ResNet18; 
Fig.~\ref{fig:adv} (b, d). When distillation is done from a transformer to a CNN (ViT $\rightarrow$ ResNet18) the fooling rates remain similar for the independent and distilled students; Fig.~\ref{fig:adv} (e). Here, we don't see the student becoming similar to the teacher to the extent observed for a CNN $\rightarrow$ CNN distillation.

%, especially if it was generated through multiple rounds of targeted attack
%The fooling method that we choose, i

\textbf{Discussion:} This result is surprising. Iterative FGSM is a white box attack, which means that it has full access to the target model (teacher), including its weights and gradients. Thus, the adversarial examples are specifically crafted to fool the teacher. In contrast, the student never has direct access to the weights and gradients of the teacher, regardless of the distillation method. Yet, by simply trying to mimic the teacher's soft probabilities ($KL$) or an intermediate feature layer
\yj{($Hint$, $CRD$)}, the student network inherits, to some extent, the particular way that a teacher is fooled. 

% But, does this happen only because the distillation process improves the student's accuracy? To test this hypothesis, we performed $VGG19 \rightarrow VGG11$ distillation using $KL$ and $Hint$ using different hyper-parameter values ($\alpha, \beta, \gamma$ in Eq.~\ref{eq:kl}, ~\ref{eq:hint}), and chose the distilled students that are no more accurate than the independent student. The top-1 accuracy of the models are: (i) $S_{Ind}$: 68.89\%, (ii) $S_{KL}$: 68.48\% and (iii) $S_{Hint}$: 67.94\%. Fig.~\ref{fig:adv} (e) shows the results of attacking these students using successful adversarial images crafted for VGG19. Interestingly, the fooling rates for the distilled students are still higher compared to the independent student. We tie this result to the question raised in the beginning: even if distillation does not improve the accuracy, as was shown in \cite{cho-iccv2019}, the distillation procedure could still be transferring some other form of knowledge into the student.  

%However, we want to make sure that the
We conduct an additional study to ensure that the reason the distilled student is getting fooled more is because it is being attacked specifically by its \emph{teacher's} adversarial images; i.e., if those images are designed for some other network, would the difference in fooling rates still be high?
%We perform an additional experiment to rule out a different conclusion.  Specifically, it may be that the distillation process itself makes the student more vulnerable to \emph{any} adversarial image, and not necessarily to those designed specifically for the teacher. 
We test this in the VGG19 $\rightarrow$ VGG11 setting. This time we generate adversarial images ($I \rightarrow I^{adv}$) to fool an ImageNet pre-trained ResNet18 network, instead of VGG19 (the teacher).  We then use those images to attack the same VGG11 students from the VGG19 $\rightarrow$ VGG11 setting.  In Fig.~\ref{fig:adv} (c), we see that the fooling rates for the independent and distilled students remain similar. This indicates that distillation itself does not make the student more vulnerable to \emph{any} adversarial attack, and instead, an increase in fooling rate can be attributed to the student's inheritance of the teacher's adversarial vulnerability.

%; i.e., there is no significant increase in fooling rate for the distilled students

%However, there could be a different conclusion drawn from this result. It is possible that the distillation process makes the student vulnerable to \emph{any} adversarial image, and not necessarily to those designed for the teacher. 

%\subsection{Does knowledge about invariance to change get distilled?}
\vspace{-2pt}
\subsection{Does invariance to data transformations get distilled?}\label{sec:inv}
\vspace{-2pt}

We have studied whether the response properties on single images get transferred from the teacher to the student. Now suppose that the teacher is invariant to certain changes in data, either learned explicitly through data augmentation or implicitly due to architectural choices. Can such properties about \emph{changes in images} get transferred during distillation? 
%have its own biases or 
%In other words, if the teacher was more invariant to a particular change, and the student less so, will it become more invariant to that change after distillation? 
%All the experiments that follow will have this theme: we create two versions of a test image, $X_1$ and $X_2$, which are different random variations of a particular change. For example, $X_1$ and $X_2$ could be the same image augmented to have different hue levels. We pass both images through a network \ut{either teacher or student} and report how frequently their predicted classes match. We call this the agreement score, which we compute over the 50k images of the ImageNet validation set.

%\subsubsection{Color invariance}\label{sec:color_inv}

\textbf{Experimental setup:} We study color invariance as a transferable property. 
%(The appendix (C.4) contains a similar experiment where we study invariance to random shifts in images instead and find a similar result).
We train three models for ImageNet classification: a teacher, a distilled and an independent student. While training the teacher, we add color jitter in addition to the standard augmentations (random crops, horizontal flips). Specifically, we alter an image's brightness, contrast, saturation, and hue, with magnitudes sampled uniformly in [0, 0.4] for the first three, and in [0, 0.2] for hue. This way, the teacher gets to see the same image with different color properties during training and can become color invariant. When training the student, we only use the standard augmentations \emph{without} color jittering, and see whether such a distilled student can indirectly inherit the color invariance property through the teacher. 

\begin{figure*}[t]
    \centering
    \includegraphics[width=1\textwidth]{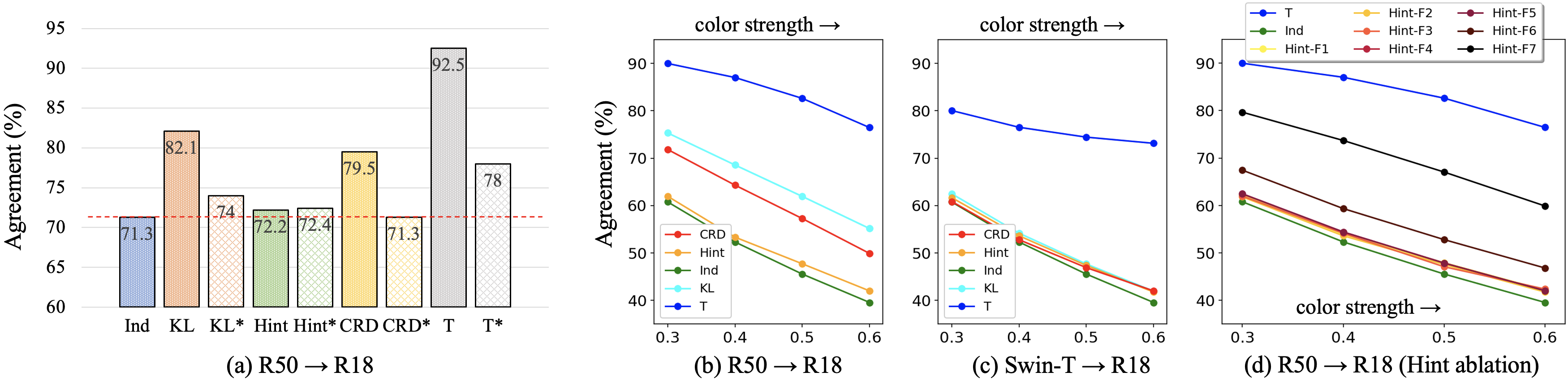}
    \vspace{-15pt}
    \caption{\textbf{(a)} Agreement between two images  with different color properties. * indicates distillation done by a teacher T$^{*}$ not trained to be color invariant. \textbf{(b, c)} Agreement between two images having increasingly different color properties. \textbf{(d)} Effect of different layers used in $Hint$ distillation.}
    \label{fig:color}
    \vspace{-5pt}
\end{figure*}
%And if the student does become invariant to color changes, then it should produce the same classification label for two different color augmentations of the same image. 
%In contrast, an independent student is less likely to become color invariant, since it never learned about it explicitly.
%inherit that property and 

\textbf{Results:} We start with the ResNet50 $\rightarrow$ ResNet18 configuration. After training, we evaluate the models on 50k ImageNet validation images. For each image $X$, we construct its augmented version $X'$ by altering the color properties in the same way as done while training the teacher. Fig.~\ref{fig:color} (a) depicts the agreement scores between $X$ and $X'$ (y-axis) for different models (x-axis). The teacher (T), being trained to have that property, achieves a high score of 92.5\%. The independent student, which is not trained to be color invariant, has a much lower score of 71.3\%. But, when the color-invariant teacher is used to perform distillation using $KL$ or $CRD$, the agreement scores of the students jump up to 82.1\% and 79.5\% respectively. To ensure that this increase is not due to some regularizing property of the distillation methods that has nothing to do with the teacher, we repeat the experiment, this time with a ResNet50 teacher (T$^{*}$) that is not trained with color augmentation. The new agreement scores for the distilled students (marked by *; e.g., KL$^{*}$) drop considerably compared to the previous case. This is a strong indication that the student does inherit \emph{teacher-specific} invariance properties during distillation.

In Fig.~\ref{fig:color}(b), we show that this trend in agreement scores (y-axis) holds even when the magnitude of change in brightness, contrast and saturation (x-axis) is increased to create $X'$ from $X$. We repeat this experiment for Swin-tiny (a transformer) \cite{swin} $\rightarrow$ ResNet18 in Fig.~\ref{fig:color}(c). We again see some improvements in the agreement scores of the distilled students. This increase, however, is not as significant as that observed in the previous CNN $\rightarrow$ CNN setting. Throughout this work, we find that distilling the properties from a transformer teacher into a CNN student is difficult. The implicit biases introduced due to architectural differences between the teacher (transformer) and student (CNN) seem too big, as was studied in \cite{cnn-vs-vit}, to be overcome by current distillation methods.

We also note the ineffectiveness of $Hint$ in distilling this knowledge. Our guess for this is the choice of $l$ for distillation, which is typically set to be in the middle of the network as opposed to deeper layers where $KL$ and $CRD$ operate.
%Our initial guess for this is because the default $Hint$ asks the student to mimic the teacher at somewhere in the middle of the network. This is different than $KL$ and $CRD$, which operate at deeper levels of the network. 
So, we perform an ablation study for $Hint$, where the student mimics a different feature of the teacher each time; starting from a coarse feature of resolution $56 \times 56$ (F1) to the output scores (logits) of the teacher network (F7). We plot the agreement scores in Fig.~\ref{fig:color}(d), where we see that the score increases as we choose deeper features. Mimicking the deeper layers likely constrains the overall student more compared to a middle layer, since in the latter case, the rest of the student (beyond middle layer) could still function differently compared to the teacher.
%However, why should the information about color invariance be better encoded in deeper rather than shallower layers is an open question.

\textbf{Discussion:} In sum, color invariance can be transferred during distillation. This is quite surprising since the teacher is not used in a way which would expose any of its invariance properties to the student. Remember that all the student has to do during distillation is to match the teacher's features for an image $X$; i.e., the student does not get to see its color augmented version, $X'$, during training. If so, then why should it get to know how the teacher would have responded to $X'$? 
%We try providing an answer in Sec.~\ref{sec:why}.  

% \begin{figure*}[t]
%     \centering
%     \includegraphics[width=1\textwidth]{figs/crop.pdf}
%     \vspace{-5pt}
%     \caption{Crop invariance}
%     \label{fig:color}
% \end{figure*}

% \subsubsection{Shift invariance}
% The $X_1$ and $X_2$ discussed in the previous section could have different aspect ratios. However, shift invariance is a more subtle property
\vspace{-2pt}
\subsection{Does knowledge about unseen domains get distilled?}\label{sec:ood}
\vspace{-2pt}

\begin{figure*}[t!]
    \centering
    \includegraphics[width=1\textwidth]{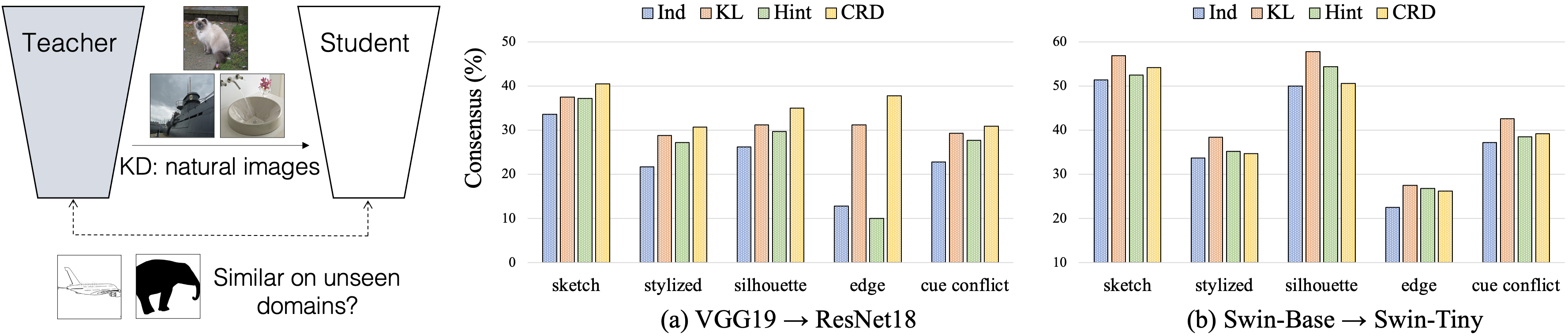}
    \vspace{-15pt}
    \caption{\textbf{Left:} Does knowledge transferred about one domain give knowledge about other unseen domains? \textbf{Right:} Consensus scores between teacher and student for images from unseen domains.}
    \label{fig:ood}
    \vspace{-5pt}
\end{figure*}

So far, our analysis has revolved around the original ImageNet dataset, something that was used to perform the distillation itself. So, does the teacher only transfer its knowledge pertaining to this domain, or also of domains it has never seen (Fig.~\ref{fig:ood} left)? 

\textbf{Experimental setup:} To test this, we first perform distillation using two settings (i) VGG19 $\rightarrow$ ResNet18 and (ii) Swin-Base \cite{swin} $\rightarrow$ Swin-Tiny, where the training of the teacher, as well as distillation is done on ImageNet. During test time, we take an image from an unseen domain and see how frequently the student's and teacher's class predictions match for it (regardless of whether the predicted label is correct or incorrect), which we call the consensus score.
% Note that this agreement score is different than the one used in Sec.~\ref{sec:inv}. 
%During the distillation process, the essential goal is to make the student mimic the teacher on some dataset of interest. Does this make the student more similar to the teacher on some other domains as well, which was not used for training? 
%Note that the goal is not to report the accuracy of either the teacher or the student for this new domain, but rather to see if their behaviour is similar, even if it is incorrect. 
For the unseen domains, we consider the five datasets proposed in \cite{geirhos2021partial}: $\texttt{sketch}$, $\texttt{stylized}$, $\texttt{edge}$, $\texttt{silhouette}$, $\texttt{cue conflict}$. Images from these domains are originally from ImageNet, but have had their properties modified. 
%E.g., $\texttt{sketch}$ contains ImageNet images converted to their sketch form, $\texttt{stylized}$ contains images which have their content preserved, but with different style. 

%\yh{I think we need to define "agreement" here, since this "agreement" is different from the previous one. Also, I still think we need to have more settings, which at least has a transformer one. Since the current transferring results are not that conclusive if we see the resnet case}
\textbf{Results:} Fig.~\ref{fig:ood} (right) shows the consensus scores between the teacher and the student (y-axis). We see a nearly consistent increase in consensus brought about by the distillation methods in both settings. The extent of this increase, however, differs among the domains. There are cases, for example, in VGG19 $\rightarrow$ ResNet18, where consensus over images from $\texttt{edge}$ domain increases over 100\% (12\% $\rightarrow$  $\ge$30\%) by some distillation methods. On the other hand, the increase can also be relatively modest in certain cases: e.g. Swin-Base $\rightarrow$ Swin-Tiny in $\texttt{stylized}$ domain. 
%Now, from Sec.~\ref{sec:ref}, remember that 

\textbf{Discussion:} \cite{stanton-neurips2021} showed that the agreement in classification between the teacher and distilled student is not much different than that to an independent student on CIFAR. We find this to be true in our ImageNet classification experiments as well. That is why the increase in agreement observed for unseen domains is surprising. After all, if the agreement is not increasing when images are from the seen domain, why should it increase when they are from an unseen domain?  It is possible that there is more scope for increase in agreement in unseen domains vs seen domain. The consensus score between teacher and independent student for the seen domain is $\ge$75\% (appendix), whereas for an unseen domain, e.g. $\texttt{sketch}$, it is $\le$40\%. So, it might not be that knowledge distillation does not work, as the authors wondered in \cite{stanton-neurips2021}, but its effect could be more prominent in certain situations.     

%These results might seem in contradiction to that presented in \cite{stanton-neurips2021}, where the authors studied agreement on the validation set of ImageNet itself (not an unseen domain), and found no significant improvement by the distilled student compared to an independent student. 
%Again, we believe that this is not a trivial result: it is not clear if matching the output spaces of two complex neural networks

\vspace{-3pt}
\subsection{Other studies}
In appendix, we study additional aspects of a model such as shape/texture bias, invariance to random crops, and find that even these obscure properties can transfer from a teacher to the student. We also explore the following questions: If we can find alternative ways of increasing a student's performance (e.g. using crafted soft labels), will that student gain similar knowledge as a distilled student? If distillation cannot increase the student's performance, is there no knowledge transferred? Finally, we also present results on some more datasets like CIFAR100~\cite{cifar100}, VLCS and PACS~\cite{li-domain}, showing that the phenomena of the implicit transfer of properties dueing knowledge distillation extends even to datasets beyond ImageNet and MNIST.   

%\yj{We might want to say what other studies we have done, and where to find them in the supp.  This can be just 2-3 lines.  For example, ``We provide more analysis on pseudo-label precision and recall, and results with true out-of-distribution unlabeled data. These can be found in the Appendix.''}

%methods trying to make the student mimic the teacher.
%that we notice throughout this work, 

%Note that the dataset used during distillation is the standard ImageNet, so any improvement observed in \ut{Fig. X} is only an indirect result of student mimicking the teacher. 
%We don't see any improvements through $Hint$, however, similar to results shown in Sec.~\ref{sec:inv}.  
\vspace{-2pt}

\vspace{-2pt}
\section{Applications}\label{sec:mnist}
\vspace{-2pt}

\begin{figure*}[t]
    \centering
    \includegraphics[width=1\textwidth]{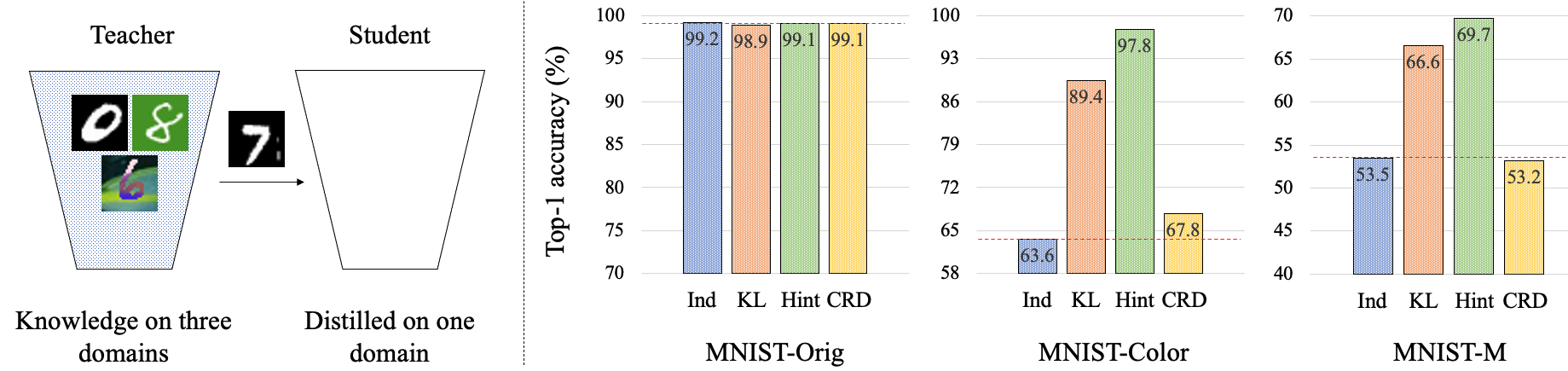}
    \vspace{-15pt}
    \caption{\textbf{Left:} The teacher is trained on three domains: $\texttt{MNIST-Orig}$, $\texttt{MNIST-Color}$, and $\texttt{MNIST-M}$. Distillation is done only on $\texttt{MNIST-Orig}$. \textbf{Right:} Test accuracy of the students. Note the increase in performance on $\texttt{MNIST-Color}$ \& $\texttt{MNIST-M}$ domains by the distilled students.}
    \label{fig:mnist}
    \vspace{-5pt}
\end{figure*}

Beyond the exploratory angle of studying whether certain properties get transferred during distillation, the idea of a student becoming similar to a teacher in a broad sense has practical implications. We discuss a good and a bad example in this section.
%Previous sections have had an exploratory angle to them, where we have primarily been interested if certain (often obscure) properties can transfer via the distillation process. Our finding, which is that the student becomes similar to the teacher on a much broader level, can have practical implications. In this section, we present two of those - a good example and a bad example.

%\ut{The previous sections provided a \yj{comprehensive study} of the indirect effects of the distillation process. While we believe that those important properties of a model, and consequently the ability to transfer them, is important in itself, their study also paves the way for two sides of practical applications - \yj{a good example and a bad example.}}
\vspace{-2pt}
\subsection{The Good: the free ability of domain adaptation}\label{sec:good_mnist}
\vspace{-2pt}
Consider the following setup: the teacher is trained for a task by observing data from multiple domains ($\mathcal{D}_1 \cup \mathcal{D}_2$). It is then used to distill knowledge into a student on only $\mathcal{D}_1$. Apart from getting knowledge about $\mathcal{D}_1$, will the student also get knowledge about $\mathcal{D}_2$ indirectly? 

%Consider the following setup: if $\texttt{T}$ is trained on $\mathcal{D}_1 \cup \mathcal{D}_2$, and is used to distill knowledge into student on $\mathcal{D}_1$, does the student's performance increase on $\mathcal{D}_2$ as well?

% Consider the following: a model ($T$) is trained for image classification on a dataset consisting of multiple domains; $\{\mathcal{D}_1 \cup \mathcal{D}_2 ... \cup \mathcal{D}_n\}$. It is then used as a teacher to distill knowledge into a student ($S_d$) on \emph{one} of those domains: $T \rightarrow S_d$ on $D_1$. The independent student ($S$) is also trained on $\mathcal{D}_1$. Will the distilled student perform better than the independent student on the rest of the domains $\mathcal{D}_2, ...\mathcal{D}_n$? 

% as our testbed
\textbf{Experimental setup:} We use MNIST digit recognition, and train the teacher on three domains: (i) $\texttt{MNIST-orig}$: original gray-scale images from MNIST, (ii) $\texttt{MNIST-Color}$: background of each image randomly colored, and (iii) $\texttt{MNIST-M}$ \cite{mnist-m}: MNIST digits pasted on random natural image patches. The student models are trained \emph{only} on $\texttt{MNIST-orig}$ and evaluated (top-1 accuracy) on all three domains. The network architecture is same for both the teacher and the student (see appendix). 

\textbf{Results:} When the independent student is trained only on $\texttt{MNIST-orig}$, its performance drops on the unseen domains, which is expected due to domain shift. The distilled students (especially $KL$ \& $Hint$), however, are able to significantly improve their performance on both unseen domains; Fig.~\ref{fig:mnist}.

%\yh{For "no color information", Do you mean the input image or the response? If it is the second one, then I feel the claim is a bit strong, since response is a function of both input and the model, which has knowledge about color} 

\textbf{Discussion:} This result shows distillation's practical benefits: once a teacher acquires an ability through computationally intensive training (e.g., training on multiple datasets), that ability can be distilled into a student, to a decent extent, through a much simpler process. The student sees the teacher's response to \emph{gray-scale} images ($\texttt{MNIST-orig}$) that lack any color information. But that information helps the student to deal better with \emph{colored} images (e.g., $\texttt{MNIST-Color}$), likely because the teacher has learned a domain-invariant representation (e.g., shape) which is distilled to the student.

\vspace{-2pt}
\subsection{The Bad: Students can inherit harmful biases from the teacher}\label{sec:bad_bias}
\vspace{-2pt}

Consider the problem of classifying gender from human faces. Imagine an independent student which performs the classification fairly across all races. The teacher, on the other hand, is biased against certain races, but is more accurate than the student \emph{on average}. Will the student, which was originally fair, become unfair after mimicking the unfair teacher?

\textbf{Experimental setup:} We consider a ResNet20 $\rightarrow$ ResNet20 setting, and use FairFace dataset \cite{fairface}, which contains images of human faces from 7 different races with their gender labeled. From its training split, we create two different subsets ($\mathcal{D}_s$ and $\mathcal{D}_t$) with the following objectives - (i) $\mathcal{D}_s$ has a particular racial composition so that a model trained on it will perform fairly across all races during test time; (ii) $\mathcal{D}_t$'s composition is intended to make the model perform unfairly for certain races. The exact composition of $\mathcal{D}_s$ and $\mathcal{D}_t$ is given in the appendix. The teacher is trained on $\mathcal{D}_t$ whereas the independent/distilled students are trained on $\mathcal{D}_s$. We use $KL$ for distillation.

\begin{wrapfigure}[11]{r}{0.5\textwidth}
	\centering
 	\vspace{-13pt}
	%\hspace*{5pt}
	\includegraphics[width=0.5\textwidth]{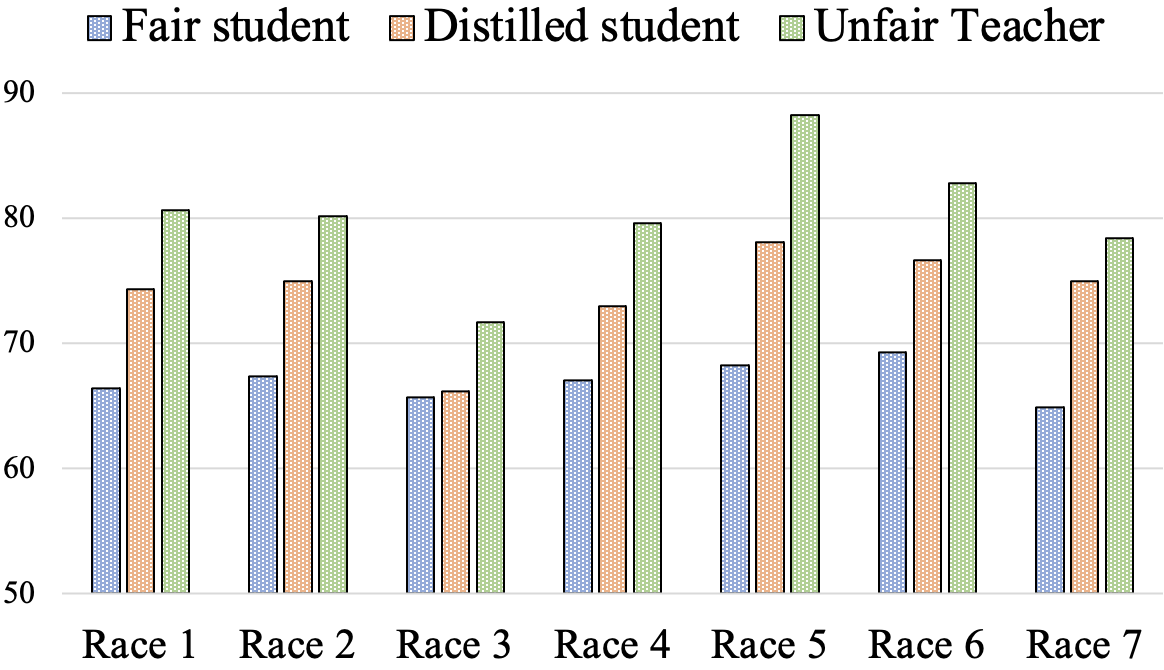}
	\vspace{-5pt}
\end{wrapfigure}
\textbf{Results:} We observe in the figure on the right that the independent student performs roughly fairly across all races, which is what we intended. The teacher is more accurate than the student, but performs relatively poorly on faces from Race 3 compared to others. After distillation, we observe that the student's \emph{overall} accuracy improves, but the gain in accuracy is less for images from Race 3. So, when it is trained on $\mathcal{D}_s$ by itself, it behaves fairly. But when it mimics the teacher on $\mathcal{D}_s$, it becomes unfair. This follows the observation made in \cite{lukasik2021}, where the authors found that the performance gains during distillation might not be spread uniformly across different sub-categories.

\textbf{Discussion:} The practical takeaway from the experiment is that knowledge distillation can bring forth behaviour which is considered socially problematic if it is viewed simply as a blackbox tool to increase a student's performance on test data, as it did in the previous example. Proper care must hence be taken regarding the transfer/amplification of unwanted biases in the student.

% \subsection{Key takeaways:}
% Add text later.

\section{Why does knowledge distillation work in this way?}
\begin{figure*}[t]
    \centering
    \includegraphics[width=1\textwidth]{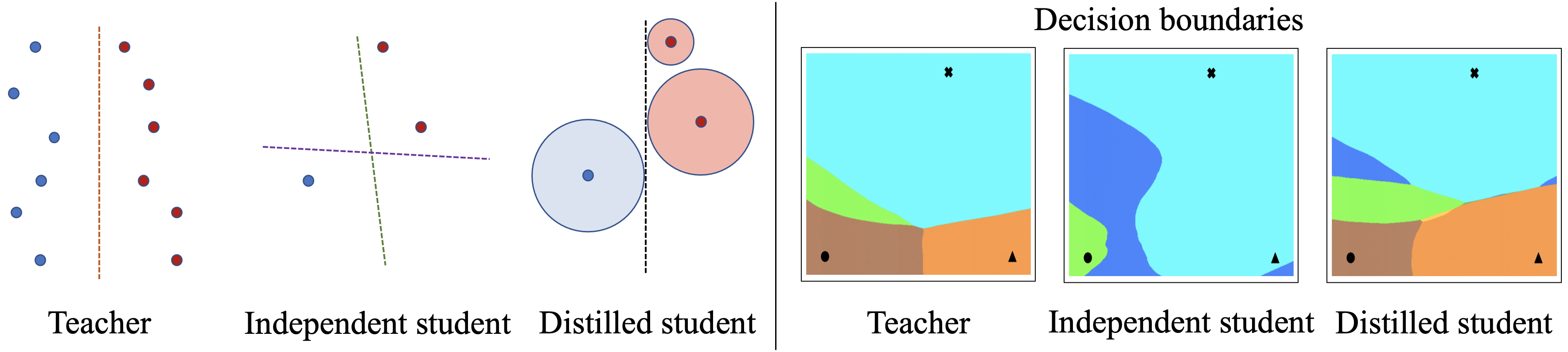}
    \vspace{-15pt}
    \caption{\textbf{Left.} Teacher learns on the the entire dataset. Independent student without access to the full dataset might learn some spurious correlations. Distillation provides constraints to reconstruct the teacher's decision boundary. \textbf{Right.} Sampled decision boundaries for points from MNIST-Color visualized using \cite{somepalli2022can}. The distilled student is better at reconstructing the teacher's decision boundary.}
    \label{fig:thoughtexp}
    \vspace{-5pt}
\end{figure*}

%Why should a teacher's response to an image contain such rich information about that teacher's implicit properties? This is what we explore in this section.
\paragraph{An Illustrative Example.} Why should a teacher's response to an image contain such rich information about its implicit properties? We first intuitively  explain this through a toy classification problem using $KL$ as the distillation objective. Fig.~\ref{fig:thoughtexp} (left) shows data points from two classes (red and blue). \ut{The teacher has access to the complete set, thereby learning to classify them appropriately (orange decision boundary). On the other hand, suppose that the training of the independent and distilled student is done on a subset of the points. The independent student can learn to classify them in different ways, where each decision boundary can be thought of as looking for different features, hence giving rise to different properties.} But the distilled student is trying to mimic the teacher's class probabilities, which reflect the distance from the (orange) decision boundary of the teacher, visualized as circles around the points. Then the decision boundary of the distilled student should remain tangential to these circles so that it preserves the teacher's distances. So, this very simple case depicts that mimicking the class probabilities can force the student's decision boundary to resemble that of the teacher, which is \emph{not} true for the independent student.
%different set of
%In order to give an answer to that question, we first explore the $KL$ objective and offer intuition in a simple situation. As logits are a measure of the Euclidean distance, we contend that the teacher's logits give information on the location of input data points relative to the teacher's decision boundary. Therefore, with this additional knowledge, the student is able to exactly identify where the teacher's decision boundary is. 
%In addition, By carefully carrying out experiments in the case of complex neural networks, we demonstrate that this is, in fact, the case.
% This in turn helps constrain the decision boundary which is learned by the student. By carefully carrying out experiments in the case of complex neural networks, we demonstrate that this is, in fact, the case.
%\subsection{Simple Case of a Linear Classifier}
%\subsubsection{Intuition}
%Consider a linear classifier as the teacher model Fig. \ref{fig:thoughtexp}(a) which has access to many training data instances and can correctly discern cats from dogs. We now fit the independent student and the distilled student to fewer data points and initialize both of them as linear classifiers. By doing this, we shed some light on the \emph{dark knowledge} the teacher's logits had to offer.
%In this example, because the independent student \ref{fig:thoughtexp}(b) is only making an attempt to distinguish between cat and dog instances, it's possible that it might end up learning the wrong hypothesis (indicated by the yellow line). Contrarily, the distilled student \ref{fig:thoughtexp}(c) makes an attempt to match the logits provided by the teacher. Since logits are a measure of euclidean distance, we argue that distillation in this case provides information about the distance of the data points from the decision boundary (represented by the circles). In order to maintain distances from the decision boundary to the data points, the student must learn a decision boundary that is tangential to each circle. Thus the distilled student—which was trained on a small sample size of data—is able to exactly reproduce the decision boundary of the teacher. Below, we formulate and mathematically verify the aforementioned case.

A mathematical formulation of the same case can be considered as follows. Let $D$ be a linearly separable dataset: $D \in \{(x_1, y_1), (x_2, y_2), ......, (x_n, y_n) | x_i \in \Re^m, y_i \in \{-1, 1\}\}$. For an arbitrary linear classifier having weights $W \in \Re^{m \times 1}$ and biases $b \in \Re$, the binary classification objective for the teacher and the independent student can be described as $\forall (x_i, y_i) \in D, (W^Tx_i + b)y_i > 0$.

% \\
%  \begin{equation} \label{eq:binary_classification}
%  \begin{split}
%  \hspace*{1cm}\forall (x_i, y_i) \in D, (W^Tx_i + b)y_i > 0
%  \end{split}
%  \end{equation}
%  \\
Since the dataset is linearly separable, both the teacher as well as the independent student can solve this system of inequalities in many ways. Let $\{W_{t}$, $b_{t}\}$ be the parameters of one particular teacher. 

Next, we look at the objective of the distilled student, who tries to map $x_i$ to $z_i$ = $W^{T}_tx_i + b_{t}$, instead of $y_i$. That is, the optimal solution satisfies: $\forall (x_i, z_i) \in D_{dist}, W_s^Tx_i + b_s = z_i$.
%To properly examine the distillation objective, we define a modified version of the previous dataset $D$ called $D_{dist}$.
%\\\\
%\hspace*{1cm}$D_{dist} \in \{(x_1, z_1), (x_2, z_2), ......, (x_n, z_n) | x_i \in \Re^m, z_i = W^{T}_t^*x_i + b_{t}^*\}$
%\\\\
%The dataset $D_{dist}$ contains the same feature vectors as the original dataset {D}. Let the distilled student have a weight and bias $W_s$ and $b_s$ such that $W_{s} \in \Re^{m \times 1}$ and $b_{s} \in \Re$. 
% \\
% \begin{equation} 
% \begin{split}
% \hspace*{1cm}\forall (x_i, z_i) \in D_{dist}, W_s^Tx_i + b_s = z_i
% \end{split}
% \label{eq:toy_dist}
% \end{equation}
% \\
% Let $X_{dist}$ represent a data matrix such that $X_{dist} \in \Re^{m \times p}$ where $p$ represents the number of data points sampled from the dataset $D_{dist}$. Equation 4 can be rewritten as,
% \begin{equation} \label{eq5}
% \begin{split}
% W_s^TX_{dist} + b_s = Z 
% \end{split}
% \end{equation}
% \\
% where,  $Z \in R^{1 \times p}$ is a matrix containing scores from the dataset.
This system of linear equations will have a unique solution provided it has access to at least $m+1$ training instances and that these instances are linearly independent. Since we know that the teacher's weights and biases, $\{W_{t}$, $b_{t}\}$, can satisfy this equation, we can conclude that if there exists a unique solution, then $W_s = W_t$ and $b_s = b_{t}$. So, the distilled student, when solving the distillation objective, will recover the weights and biases of the teacher, and consequently its decision boundary.

% \begin{equation} \label{eq6}
% \begin{split}
% W_s = W^{T}_t^*
% \end{split}
% \end{equation}
% \begin{equation} \label{eq7}
% \begin{split}
% b_s = b_{t}^*
% \end{split}
% \end{equation}
% \\
%\as{Explain the information loss perspective and explain why decision boundary reconstruction is not trivial}
\paragraph{Complex neural networks:} Given that the distances of data points, i.e., decision boundary, can be preserved by the distilled student in the very simple case presented above, we now study whether they can be preserved in a more complex setting involving multi-layer neural networks as well. 

%our hypothesis is that the distilled student inherits the decision boundary from the teacher, and thus absorbs other properties of the teacher as well. 
%if the decision boundary is inherited from the teacher, the student will absorb other teacher's properties as well. 
%However, this setting is fundamentally different from that of the simple linear case. Studies have revealed that the information about the original data points is only barely conserved in activations at the deeper layers \cite{shwartzziv2017opening}.
There have been works \cite{goldblum2020adversarially, somepalli2022can} which have studied this for the $KL$ objective: if the teacher and the students are all trained on the same dataset $\mathcal{D}$, the decision boundary of the teacher will be more similar to the distilled student than the independent student \emph{for points from} $\mathcal{D}$. And while this result is useful, it is not clear whether (i) this can help explain why the student inherits teacher's knowledge about other domains, and (ii) whether this holds true for distillation objectives other than $KL$.

To study this, we revisit the setup of Sec.~\ref{sec:good_mnist}, where the teacher was trained on three domains and distillation was performed on one (Fig.~\ref{fig:mnist} left). We study what happens to the decision boundary similarity between the students and the teacher on all three domains. To compute that similarity between two neural networks, we use the method introduced in \cite{somepalli2022can}. We randomly sample a triplet of data points from a given domain and create a plane that passes over that triplet. The networks that are being compared receive the same dataset of points in this plane, and we compute intersection over union between their predictions. We report the score averaged over 50 triplets.

\begin{table}[!htb]

\centering
\begin{tabular}{lllll}

            & \textbf{Ind} & \textbf{KL} & \textbf{Hint} & \textbf{CRD} \\
            \hline
$\texttt{MNIST-Orig}$  &0.9107 & 0.9341 & 0.9317 &  0.9062   \\

$\texttt{MNIST-Color}$ &0.4156 & 0.6711 & 0.7980 &  0.4733   \\

$\texttt{MNIST-M}$     &0.4691 & 0.5872 & 0.6541 &  0.4862  \\
\hline
\end{tabular}
\caption{Decision boundary similarity calculated between students and the teacher.}
\label{Tab:mnist}
% \begin{tabular}{lllll}

%             & \textbf{Ind} & \textbf{KL} & \textbf{Hint} & \textbf{CRD} \\
%             \hline
% $\texttt{MNIST-Orig}$  &0.8746 & 0.9444 & 0.9281 & 0.8850 \\

% $\texttt{MNIST-Color}$ &0.3839 & 0.8763 & 0.8119 & 0.4065    \\

% $\texttt{MNIST-M}$     &0.8831 & 0.9350 & 0.9377 & 0.8837   \\
% \hline
% \end{tabular}
% \caption{Decision Boundary Similarity calculated between networks trained with various distillation objectives on $\texttt{MNIST-M}$ and the Teacher.}
% \label{Tab:mnistm}
\end{table}

%\subsubsection{Does Knowledge Distillation reconstruct the teacher's decision boundary for even hidden domains?}
%In order to determine if the three distillation techniques result in the reconstruction of the decision boundary in each of the three domains, we train the student solely on $\texttt{MNIST-orig}$ and compute the decision boundary similarity between the teacher and the student on all three domains. We compare this with the similarity of the teacher and the independent student.

\textbf{Results:} Table \ref{Tab:mnist} shows the similarity scores, where we see that the distilled students' decision boundary are almost always more similar to the teacher's than the independent student's. This is particularly evident when the scores are computed for domains not seen by the students; e.g., for $\texttt{MNIST-Color}$, the similarity score increases from 0.416 to 0.671 for the student distilled with $KL$.  
%Our research demonstrates that there is a lot more similarity between the teacher's decision boundary and the distilled student's decision boundary than there is between the independent student's decision boundary and the teacher's decision boundary.

\textbf{Discussion:} First, we can draw an analogy of this experimental setup with the toy example discussed before. The three $\texttt{MNIST}$ related domains (for training the teacher) are similar to the \emph{overall} set of blue+red points used to train the toy teacher (Fig.~\ref{fig:thoughtexp} left). The singular $\texttt{MNIST-orig}$ domain (for training the students) is similar to the three points available to train the toy students. \ut{Now, what the results from Table \ref{Tab:mnist} show is that the data points available for distillation can help determine the teacher's decision boundary even for points which the student did not have access to}. So, similar to how the decision boundary estimated by the toy student can correctly estimate the distances of missing blue/red points, the neural network based student can also estimate the distances of points from unseen domains, e.g., $\texttt{MNIST-color}$, in a similar way as its teacher, thereby inheriting the teacher's behavior on these points. An example of this is given in Fig.~\ref{fig:thoughtexp} (right). 

%The results indicate that if the data points available for distillation ($\texttt{MNIST-orig}$ in this case) are good enough, they will determine the true decision boundary with more certainty.   
%Distillation objectives constrain the decision boundary on the training domain, and this in turn constrains the decision boundary to an extent on the hidden domains. Furthermore, we observe that using the $Hint$ objective results in a weaker reconstruction in comparison to the other objectives. We reason that this is because, as we have shown in earlier sections, replicating the later layers brings better constraints on the decision boundary.

% DON't think we will have space to include this.

% \subsection{Contributions}
% We summarize our key understanding about knowledge distillation below,

% \begin{itemize}
%   \item We show that Knowledge Distillation can help the student inherit properties of the teacher without directly being asked to do so. 
%   \item We provide an intuition that the relative positions of the data points with respect to the decision boundary are transferred using current distillation techniques in the form of euclidean distances. We extend this hypothesis and conduct experiments to demonstrate that it holds true for complex neural networks in a variety of distillation settings.

% \end{itemize}

\section{Limitations}
While we have tried to study the distillation process in many settings, there do exist many more which remain unexplored. This is because the analysis conducted in our work has many axes of generalization: whether these conclusions hold (i) for other datasets (e.g. SUN scene classification dataset), (ii) in other kinds of tasks (object detection, semantic segmentation), (iii) in other kinds of architectures (modern CNNs, like ResNext), (iv) or for more recent distillation objectives, etc. Moreover, in Sec. 3.5 in the supplementary, we explore the transferability of shape/texture bias from teacher to student, and find some discrepancy in different distillation objective's abilities. Therefore, the conclusions drawn from our work will be more helpful if one can study all the combinations of these factors. Due to limitations in terms of space and computational requirements, however, we have not been able to study all of those combinations.  

\vspace{-5pt}
\section{Conclusion}\label{sec:conclusion}
\vspace{-5pt}
Knowledge distillation is a beautiful concept, but its success i.e, increase in student's accuracy, has often been explained by a transfer of \emph{dark knowledge} from the teacher to the student. In this work, we have tried to shed some light on this dark knowledge. 
%Our work is centered around understanding what knowledge gets transferred across different architectures and algorithms. And while we have tried to give some intuition as to why, for example, mimicking the class probabilities can give information about other properties of the teacher, we cannot generalize it to \ut{all kinds of teacher's knowledge}. 
There are, however, additional open questions, which we did not tackle in this work: given the architectures of the teacher and student, is there a limit on how much knowledge can be transferred (e.g., issues with ViT $\rightarrow$ CNN)? If one wants to actively avoid transferring a certain property of the teacher into a student (Sec.~\ref{sec:bad_bias}), but wants to distill other useful properties, can we design an algorithm tailored for that? 
%Overall, we were surprised that the distillation process can transfer properties \emph{beyond} what one is optimizing for (e.g., class probabilities). However, it is also possible for one to be surprised at the results from an opposite point of view. If viewed as model compression, one might expect knowledge distillation to create a mirror version of the teacher in a student. With this view, for example, the results in Fig.~\ref{fig:ood} (a) could be surprising because the teacher and students still agree in only $\sim$30\% of images from the $\texttt{silhouette}$ domain.
We hope this work also motivates other forms of investigation to have an even better understanding of the distillation process.

\section*{Acknowledgement}
This work was supported in part by NSF CAREER IIS2150012, and Institute of Information \& communications Technology Planning \& Evaluation(IITP) grant funded by the Korea government(MSIT) (No. 2022-0-00871, Development of AI Autonomy and Knowledge Enhancement for AI Agent Collaboration), Air Force Grant FA9550-18-1-0166, the National Science Foundation (NSF) Grants 2008559-IIS, 2023239-DMS, and CCF-2046710.

{
\bibliographystyle{plain}
\bibliography{main}

\begin{thebibliography}{10}

\bibitem{beyer-arxiv2021}
Lucas Beyer, Xiaohua Zhai, Amélie Royer, Larisa Markeeva, Rohan Anil, and Alexander Kolesnikov.
\newblock Knowledge distillation: A good teacher is patient and consistent.
\newblock In {\em arXiv}, 2021.

\bibitem{bucila-kdd2006}
Cristian Bucila, Rich Caruana, and Alexandru Niculescu-Mizil.
\newblock Model compression.
\newblock In {\em SIGKDD}, 2006.

\bibitem{cho-iccv2019}
Jang~Hyun Cho and Bharath Hariharan.
\newblock On the efficacy of knowledge distillation.
\newblock In {\em ICCV}, 2019.

\bibitem{vit}
Alexey Dosovitskiy, Lucas Beyer, Alexander Kolesnikov, Dirk Weissenborn, Xiaohua Zhai, Thomas Unterthiner, Mostafa Dehghani, Matthias Minderer, Georg Heigold, Sylvain Gelly, Jakob Uszkoreit, and Neil Houlsby.
\newblock An image is worth 16x16 words: Transformers for image recognition at scale.
\newblock In {\em arXiv}, 2020.

\bibitem{born-again}
Tommaso Furlanello, Zachary~C. Lipton, Michael Tschannen, Laurent Itti, and Anima Anandkumar.
\newblock Born again neural networks.
\newblock In {\em ICML}, 2018.

\bibitem{mnist-m}
Yaroslav Ganin, Evgeniya Ustinova, Hana Ajakan, Pascal Germain, Hugo Larochelle, Fran{\c{c}}ois Laviolette, Mario March, and Victor Lempitsky.
\newblock Domain-adversarial training of neural networks.
\newblock {\em JMLR}, 2016.

\bibitem{geirhos2021partial}
Robert Geirhos, Kantharaju Narayanappa, Benjamin Mitzkus, Tizian Thieringer, Matthias Bethge, Felix~A Wichmann, and Wieland Brendel.
\newblock Partial success in closing the gap between human and machine vision.
\newblock In {\em NeurIPS}, 2021.

\bibitem{cnn-texture}
Robert Geirhos, Patricia Rubisch, Claudio Michaelis, Matthias Bethge, Felix~A Wichmann, and Wieland Brendel.
\newblock Imagenet-trained cnns are biased towards texture; increasing shape bias improves accuracy and robustness.
\newblock In {\em ICLR}, 2019.

\bibitem{goldblum2020adversarially}
Micah Goldblum, Liam Fowl, Soheil Feizi, and Tom Goldstein.
\newblock Adversarially robust distillation.
\newblock In {\em Proceedings of the AAAI Conference on Artificial Intelligence}, 2020.

\bibitem{fgsm}
Ian Goodfellow, Jon Shlens, and Christian Szegedy.
\newblock Explaining and harnessing adversarial examples.
\newblock In {\em ICLR}, 2014.

\bibitem{gou-arxiv2021}
Jianping Gou, Baosheng Yu, Stephen~J. Maybank, and Dacheng Tao.
\newblock Knowledge distillation: A survey.
\newblock In {\em arXiv}, 2021.

\bibitem{dark_knowledge}
Geoffrey Hinton, Oriol Vinyals, and Jeff Dean.
\newblock Dark knowledge.
\newblock {\em TTIC Distinguished lecture series}, 2014.

\bibitem{hinton-neurips2014}
Geoffrey Hinton, Oriol Vinyals, and Jeff Dean.
\newblock Distilling the knowledge in a neural network.
\newblock In {\em NeurIPS Deep Learning Workshop}, 2014.

\bibitem{huang-arxiv2017}
Zehao Huang and Naiyan Wang.
\newblock Like what you like: Knowledge distill via neuron selectivity transfer.
\newblock In {\em arXiv}, 2017.

\bibitem{menon-icml}
Aditya K~Menon, Ankit~Singh Rawat, Sashank Reddi, Seungyeon Kim, and Sanjiv Kumar.
\newblock A statistical perspective on distillation.
\newblock In {\em ICML}, 2021.

\bibitem{fairface}
Kimmo Karkkainen and Jungseock Joo.
\newblock Fairface: Face attribute dataset for balanced race, gender, and age for bias measurement and mitigation.
\newblock In {\em WACV}, 2021.

\bibitem{cka}
Simon Kornblith, Mohammad Norouzi, Honglak Lee, and Geoffrey Hinton.
\newblock Similarity of neural network representations revisited.
\newblock In {\em ICML}, 2019.

\bibitem{cifar100}
Alex Krizhevsky and Geoffrey Hinton.
\newblock Learning multiple layers of features from tiny images.
\newblock In {\em Technical report, Citeseer}, 2009.

\bibitem{iter-fgsm}
Alexey Kurakin, Ian Goodfellow, and Sammy Bengio.
\newblock Adversarial machine learning at scale.
\newblock In {\em arXiv}, 2016.

\bibitem{li-domain}
Da~Li, Yongxin Yang, Yi-Zhe Song, and Timothy~M. Hospedales.
\newblock Deeper, broader and artier domain generalization.
\newblock In {\em ICCV}, 2017.

\bibitem{kd_soft_reg}
Yuan Li, Francis E.H.Tay, Guilin Li, Tao Wang, and Jiashi Feng.
\newblock Revisiting knowledge distillation via label smoothing regularization.
\newblock In {\em CVPR}, 2020.

\bibitem{swin}
Ze~Liu, Yutong Lin, Yue Cao, Han Hu, Yixuan Wei, Zheng Zhang, Stephen Lin, and Baining Guo.
\newblock Swin transformer: Hierarchical vision transformer using shifted windows.
\newblock In {\em ICCV}, 2021.

\bibitem{lukasik2021}
Michal Lukasik, Srinadh Bhojanapalli, Aditya~Krishna Menon, and Sanjiv Kumar.
\newblock Teacher's pet: understanding and mitigating biases in distillation.
\newblock In {\em arXiv}, 2021.

\bibitem{muller-neurips2019}
Rafael Müller, Simon Kornblith, and Geoffrey Hinton.
\newblock When does label smoothing help?
\newblock In {\em NeurIPS}, 2019.

\bibitem{rkd-cvpr2019}
Wonpyo Park, Dongju Kim, Yan Lu, and Minsu Cho.
\newblock Relational knowledge distillation.
\newblock In {\em CVPR}, 2019.

\bibitem{peng-iccv2019}
Baoyun Peng, Xiao Jin, Jiaheng Liu, Shunfeng Zhou, Yichao Wu, Yu~Liu, Dongsheng Li, and Zhaoning Zhang.
\newblock Correlation congruence for knowledge distillation.
\newblock In {\em ICCV}, 2019.

\bibitem{understand_kd}
Max Phuong and Christoph~H. Lampert.
\newblock Towards understanding knowledge distillation.
\newblock In {\em ICML}, 2019.

\bibitem{cnn-vs-vit}
Maithra Raghu, Thomas Unterthiner, Simon Kornblith, Chiyuan Zhang, and Alexey Dosovitskiy.
\newblock Do vision transformers see like convolutional neural networks?
\newblock In {\em NeurIPS}, 2021.

\bibitem{romero-iclr2015}
Adriana Romero, Nicholas Ballas, Samira~Ebrahimi Kahau, Antoine Chassang, Carlo Gatta, and Yoshua Bengio.
\newblock Fitnets: Hints for thin deep nets.
\newblock In {\em ICLR}, 2015.

\bibitem{grad-cam}
Ramprasaath~R. Selvaraju, Michael Cogswell, Abhishek Das, Ramakrishna Vedantam, Devi Parikh, and Dhruv Batra.
\newblock Grad-cam: Visual explanations from deep networks via gradient-based localization.
\newblock In {\em IJCV}, 2019.

\bibitem{somepalli2022can}
Gowthami Somepalli, Liam Fowl, Arpit Bansal, Ping Yeh-Chiang, Yehuda Dar, Richard Baraniuk, Micah Goldblum, and Tom Goldstein.
\newblock Can neural nets learn the same model twice? investigating reproducibility and double descent from the decision boundary perspective.
\newblock {\em arXiv preprint arXiv:2203.08124}, 2022.

\bibitem{stanton-neurips2021}
Samuel Stanton, Pavel Izmailov, Polina Kirichenko, Alexander~A. Alemi, and Andrew~Gordon Wilson.
\newblock Does knowledge distillation really work?
\newblock In {\em NeurIPS}, 2021.

\bibitem{szegedy-cvpr2016}
Christian Szegedy, Vincent Vanhoucke, Sergey Ioffe, Jonathon Shlens, and Zbigniew Wojna.
\newblock Rethinking the inception architecture for computer vision.
\newblock In {\em CVPR}, 2016.

\bibitem{tian-iclr2020}
Yonglong Tian, Dilip Krishnan, and Phillip Isola.
\newblock Contrastive representation distillation.
\newblock In {\em ICLR}, 2020.

\bibitem{tung-iccv2019}
Frederick Tung and Greg Mori.
\newblock Similarity-preserving knowledge distillation.
\newblock In {\em ICCV}, 2019.

\bibitem{fsp-cvpr2017}
Junho Yim, Donggyu Joo, Jihoon Bae, and Junmo Kim.
\newblock A gift from knowledge distillation: Fast optimization, network minimization and transfer learning.
\newblock In {\em CVPR}, 2017.

\bibitem{at-iclr2017}
Sergey Zagoruyko and Nikos Komodakis.
\newblock Paying more attention to attention: Improving the performance of convolutional neural networks via attention transfer.
\newblock In {\em ICLR}, 2017.

\bibitem{zhang2019shiftinvar}
Richard Zhang.
\newblock Making convolutional networks shift-invariant again.
\newblock In {\em ICML}, 2019.

\bibitem{dml}
Ying Zhang, Tao Xiang, Timothy~M. Hospedales, and Huchuan Lu.
\newblock Deep mutual learning.
\newblock In {\em CVPR}, 2018.

\bibitem{self_kd_soft}
Zhilu Zhang and Mert Sabuncu.
\newblock Self-distillation as instance-specific label smoothing.
\newblock In H.~Larochelle, M.~Ranzato, R.~Hadsell, M.F. Balcan, and H.~Lin, editors, {\em Advances in Neural Information Processing Systems}, volume~33, pages 2184--2195. Curran Associates, Inc., 2020.

\end{thebibliography}
}

\newpage

\section*{Appendix}

This document provides additional information complementing the main paper. First, we describe details pertaining to different distillation procedures used in Sec.~\ref{sec:train_details}. Then, in Sec.~\ref{sec:fgsm_process}, we detail the iterative FGSM \cite{iter-fgsm} used to create adversarial images. Following that, in Sec.~\ref{sec:new_analyses}, we perform more analyses to further dissect the distillation process, which corroborates our findings presented in the main paper. Finally, we present the top-1 accuracy of all the models, as well as the results shown in the main paper with their error bars, in Sec.~\ref{sec:quant_res}. Additionally, we have provided scripts used for evaluation performed in Sec. 4.2 and 4.3; please see $\texttt{readme.txt}$. 

\section{Training details}\label{sec:train_details}
\textbf{ImageNet experiments:} We first describe the hyper-parameters used for different distillation objectives.
\begin{itemize}
    \item ResNet50 $\rightarrow$ ResNet18: 
    \begin{itemize}
        \item KL: $\gamma=0.5$, $\alpha=0.5$
        \item Hint: $\gamma=1.0$, $\beta=5.0$
        \item CRD: $\gamma=1.0$, $\beta=0.8$
    \end{itemize}
    \item VGG19 $\rightarrow$ VGG11:
    \begin{itemize}
        \item KL:  $\gamma=1.0$, $\alpha=0.2$
        \item Hint: $\gamma=1$, $\beta=0.5$
        \item CRD: $\gamma=1$, $\beta=0.8$
    \end{itemize}
    \item VGG19 $\rightarrow$ ResNet18:
    \begin{itemize}
        \item KL: $\gamma=0.9$, $\alpha=0.1$
        \item Hint: $\gamma=1$, $\beta=0.2$
        \item CRD: $\gamma=1$, $\beta=1.2$
    \end{itemize}
    \item ViT $\rightarrow$ ResNet18:
    \begin{itemize}
        \item KL: $\gamma=1.0$, $\alpha=0.2$
        \item Hint: $\gamma=1$, $\beta=1$
        \item CRD: $\gamma=1$, $\beta=0.2$
    \end{itemize}
    \item Swin-Base $\rightarrow$ Swin-Tiny:
    \begin{itemize}
        \item KL: $\gamma=0.1$, $\alpha=0.9$
        \item Hint:  $\gamma=1$, $\beta=1$
        \item CRD:  $\gamma=1$, $\beta=0.8$
    \end{itemize}
    \item ResNet50 (sty) $\rightarrow$ ResNet18:
    \begin{itemize}
        \item KL (lower): $\gamma=0.1$, $\alpha=0.9$
        \item KL (higher): $\gamma=0.9$, $\alpha=0.1$
        \item Hint (lower): $\gamma=1.0$, $\beta=0.2$
        \item Hint (higher): $\gamma=1.0$, $\beta=100.0$
        \item CRD (lower): $\gamma=1.0$, $\beta=0.8$
        \item CRD (higher): $\gamma=1.0$, $\beta=1.2$
    \end{itemize}
    \item ResNet50 (col) $\rightarrow$ ResNet18:
    \begin{itemize}
        \item KL: $\gamma=0.5$, $\alpha=0.5$
        \item Hint: $\gamma=1.0$, $\beta=5.0$
        \item CRD: $\gamma=1.0$, $\beta=0.8$
    \end{itemize}
    \item ResNet50 $\rightarrow$ ResNet18 (w/o crop):
    \begin{itemize}
        \item KL: $\gamma=0.5$, $\alpha=0.5$
        \item Hint: $\gamma=1.0$, $\beta=0.2$
        \item CRD: $\gamma=1.0$, $\beta=0.8$
    \end{itemize}
\end{itemize}

The temperature used in $KL$ (Eq. 1 in main paper) is set to 4, and the temperature used in $CRD$ (Eq. 3 in main paper) is set to 0.07. For CRD, the number of negative samples ($N$ in Eq. 3) is set to 16384. For the other details, we follow the official PyTorch recommendations for training CNN-based classification models on ImageNet.\footnote{Link can be found \href{https://github.com/pytorch/examples/tree/main/imagenet}{here}.} We train the independent students for 90 epochs, and all the distilled students for 100 epochs on ImageNet. For teacher models, we try to use those officially provided by PyTorch, whenever available. For all CNN teachers (except for stylized Res50 which is taken from here\footnote{Stylized Res50 can be found \href{https://github.com/rgeirhos/texture-vs-shape}{here}.}) and ViT, we take models from PyTorch torchvision model zoo.\footnote{Link can be found \href{https://pytorch.og/vision/stable/models.html}{here}.} For Swin transformer models, we follow the training process and pretrained models given by the authors.\footnote{Swin training code and teacher models are taken from \href{https://github.com/microsoft/Swin-Transformer}{here}.} We use one 3090 Ti for training ResNet18, and two 3090 Ti for training VGG11. Each experiment takes about 2-3 days. Four A6000 are used to train Swin-T, which takes around 5 days to train. 

When performing distillation using $Hint$, we need to specify the intermediate layers at which the student will mimic the teacher. Following~\cite{romero-iclr2015}, we usually choose layers in the middle for that purpose. For ResNets, we choose feature after the second residual block, which has a resolution of 28$\times$28. For VGG11 and VGG19, we choose feature after 4th and 7th conv layer whose resolution is 56$\times$56. For Swin, we choose the feature coming after `stage 2' (refer to Fig3 in~\cite{swin}), which produces a feature of 28$\times$28 resolution. In the case of ViT-B-32 $\rightarrow$ ResNet18, the intermediate layer for ResNet18 is chosen after the fourth residual block (right before average pooling), which produces a feature of 7$\times$7 resolution. For ViT-B-32, we choose the last layer of the encoder backbone (right before classification head), which outputs a feature having 50 dimensions. Here, we remove the classification token feature and reshape the rest into a 7$\times$7 representation.      

Note that (i) ResNet50 (sty) denotes the ResNet50 teacher trained on Stylized ImageNet dataset, which is used in Section 4.5 in the main paper; (ii) ResNet50 (col) denotes the ResNet50 teacher trained with additional color augmentations, used in Section 4.3 (color-invariance experiment); (iii) ResNet18 (w/o crop) denotes the students trained without crop augmentations used in Section 4.3 (crop-invariance experiment). Finally, the further bifurcation in ResNet50 (sty) $\rightarrow$ ResNet18 i.e., lower vs higher, denotes the hyper-parameters used when we put a lower vs higher weight on the distillation loss component, relative to the cross-entropy loss.

\textbf{MNIST experiments:} The architecture of both the teacher and the student, as well as all the other training details (e.g. batch size, learning rate) is taken from the standard example given by PyTorch: $\texttt{Conv(32)} \rightarrow \texttt{ReLU} \rightarrow \texttt{Conv(64)} \rightarrow \texttt{ReLU} \rightarrow \texttt{MaxPool(2)} \rightarrow \texttt{dropout(0.25)} \rightarrow \texttt{Linear(9216, 128)} \rightarrow \texttt{ReLU} \rightarrow \texttt{dropout(0.5)} \rightarrow \texttt{Linear(128, 10)}$.\footnote{Network's architecture can be found \href{https://github.com/pytorch/examples/blob/main/mnist/main.py}{here}.}
The distillation specific hyper-parameters are listed below:
\begin{itemize}
    \item $KL$: $\gamma=0.1$, $\alpha=0.9$, $\tau=8$
    \item $Hint$: $\gamma=1.0$, $\beta=2.0$, $\texttt{Conv(64)}$ is chosen as the intermediate layer for both the teacher and the student.
    \item $CRD$: $\gamma=1.0$, $\beta=0.1$, $\tau=0.1$, no. of negative samples ($N$) = 32.
\end{itemize}
%For all the experiments, we run two different seeds: seed 0 and seed 1.  

% : we train the models using Adam optimizer~\cite{kingma2019adam}, a batch size of 256, where initial learning rate is 0.1 (for ResNet) and 0.01 (for VGG) with decay for every 30 epoches, and training is done for \ut{90} epochs for independent and \ut{100} epochs for distilled students.

\section{Process of creating the adversarial images}\label{sec:fgsm_process}
In Section 4.2 of the main paper, we  mentioned using Iterative-FGSM \cite{fgsm, iter-fgsm} for converting a clean image ($I$) to its adversarial form ($I^{adv}$). Here, we describe that conversion process in detail. First, we pass the clean image through the target network (to be fooled). Then we compute the gradient of the loss function with respect to the image ($\nabla_I$), and then update the image in the \emph{opposite} way, so as to maximize the loss ($J(I, y_{true})$). The update is bounded to be within a range [$I - \epsilon, I + \epsilon$], so that the change in the image is imperceptible. This whole process constitutes one step of FGSM, and the iterative version of this method does this for $k$ steps ($k=5$ in our case). The process can be depicted formally through Eq.~\ref{eq:fgsm}, where $\alpha$ controls the step size:
\begin{align}
    I^{adv}_{0} = I, \quad 
    I^{adv}_{t+1} = Clip_{I, \epsilon}\bigl\{ I^{adv}_{t} + \alpha \sign \bigl( \nabla_X J(I^{adv}_{N}, y_{true})  \bigr) \bigr\} 
\label{eq:fgsm}
\end{align}

\begin{figure*}[t]
    \centering
    \includegraphics[width=1\textwidth]{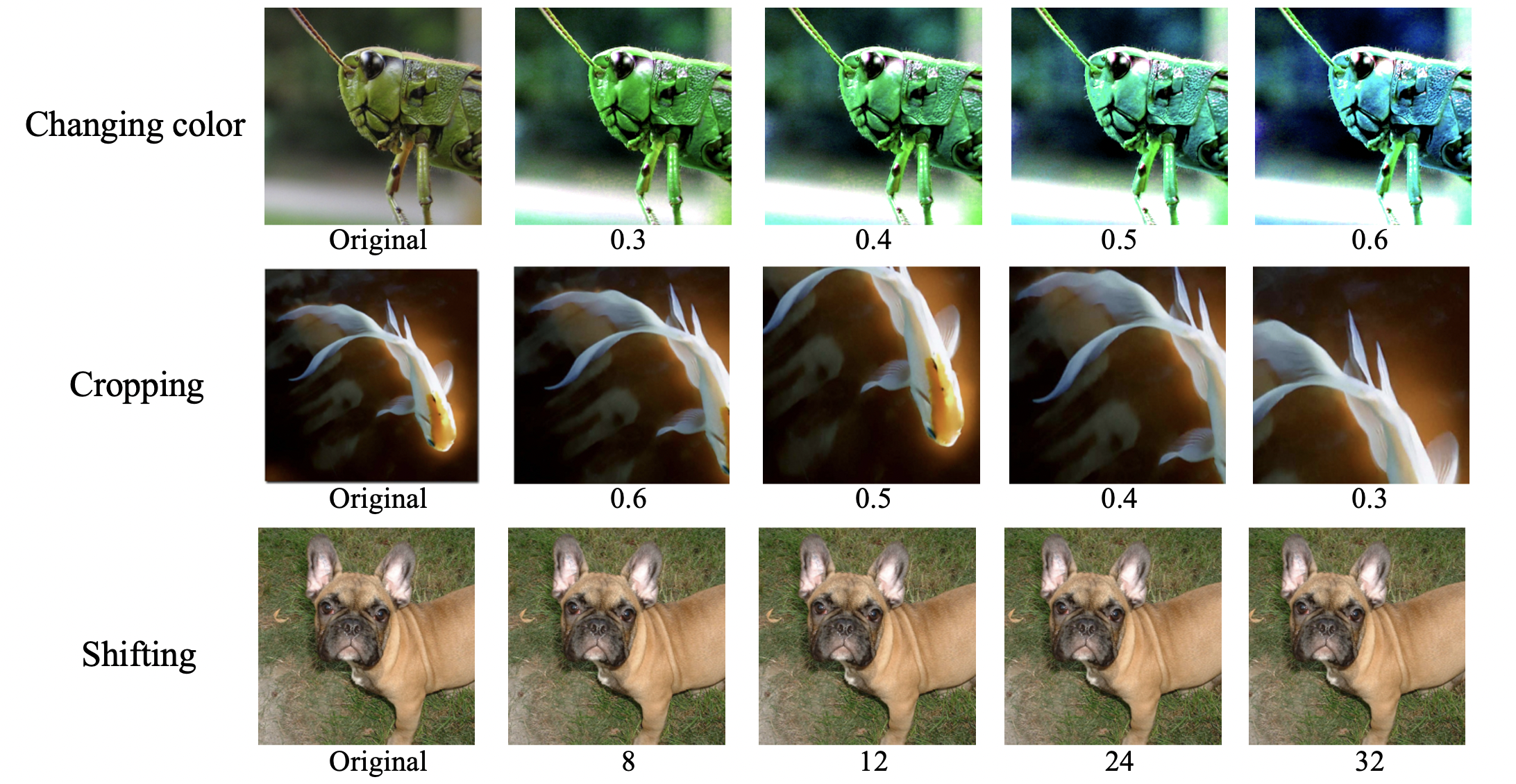}
    % \vspace{-13pt}
    \caption{ Visualizing the effect of data transformations. \textbf{Top:} Altering the color properties of an image (original) with increasing strengths. \textbf{Middle:} Taking random crops of an image (original) with different scale size. \textbf{Bottom:} Shifting the image left by different amounts. Color/crop invariance is studied in Sec. 4.3 of the main paper, and shift invariance is studied in Sec.~\ref{sec:shift_inv}.} 
    \label{fig:aug_vis}
    % \vspace{-5pt}
\end{figure*}

\begin{figure*}[t!]
    \centering
    \includegraphics[width=1\textwidth]{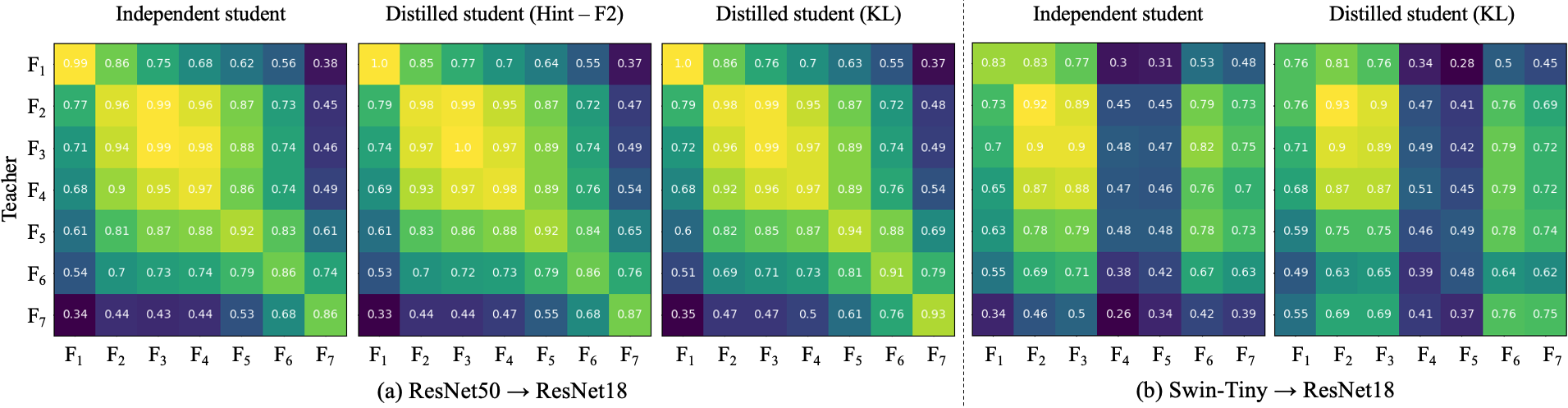}
    \vspace{-15pt}
    \caption{Centered kernel alignment (CKA) scores for various distillation settings. \textbf{Left:} Comparison of the teacher's representations with the independent and two distilled students ($KL$ and $Hint$). \textbf{Right:} Comparison of the teacher (Swin-Tiny) with independent and distilled student ($KL$).}
    \label{fig:cka}
    \vspace{-10pt}
\end{figure*}

\section{More analyses}\label{sec:new_analyses}

\subsection{Can distillation work even without increasing student's performance?}\label{sec:no_inc}

\begin{wrapfigure}[12]{r}{0.35\textwidth}
	\centering
 	\vspace{-5pt}
	%\hspace*{5pt}
	\includegraphics[width=0.35\textwidth]{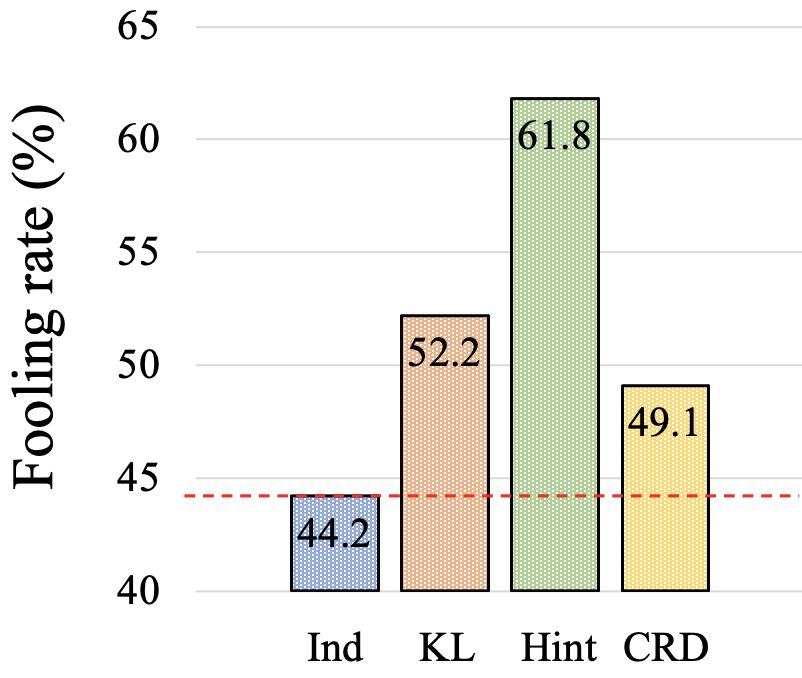}
	\vspace{-5pt}
\end{wrapfigure}
In the experiments discussed in the main paper, the distillation objective increases the performance of the student, compared to an independent student. However, it is possible that this does not happen, as was discussed in \cite{cho-iccv2019}. What do we conclude from that phenomenon? Is it that there is no knowledge transferred from the teacher to the student? In this section, we discuss such scenarios. We perform ResNet50 $\rightarrow$ ResNet18 distillation using all the distillation methods, using different hyper-parameter values ($\alpha, \beta, \gamma$ in Equation 1 and 2 in main paper), and choose the distilled students that are no more accurate than the independent student.  The top-1 accuracy of the models are: (i) $S_{Ind}$: 70.03\%, (ii) $S_{KL}$: 69.23\%, (iii) $S_{Hint}$: 70.05\% and (iv) $S_{CRD}$: 69.79\%. Figure on the top shows the results of attacking these students using successful adversarial images crafted for ResNet50. Interestingly, the fooling rates for the distilled students are still higher compared to the independent student. So, while judging a distillation setup based on the increase in student's performance is fair, it is \emph{not} that the knowledge distillation does not work if the student's performance is not increasing.

% \begin{figure*}[t]
%     \centering
%     \includegraphics[width=0.9\textwidth]{supp_fig/shift_softlabel.png}
%     % \vspace{-15pt}
%     \caption{ \textbf{(a)} Agreement between two images having increasingly shifting pixel distance. \textbf{(b)} Color invariance comparison between students distilled from a 'true' teacher and a 'dummy' teacher.  }
%     \label{fig:shift_softlabel}
%     \vspace{-5pt}
% \end{figure*}

\subsection{Can \emph{any} soft label transfer a similar knowledge?}\label{sec:soft_label}
When performing distillation through $KL$, the student has an additional target of \emph{soft labels} from the teacher to match. In another line of work on `label smoothing', converting the one-hot ground truth label into a softer version has also shown to improve a model's test performance \cite{szegedy-cvpr2016, muller-neurips2019, kd_soft_reg, self_kd_soft}. Could this mean that using any soft label, and not necessarily obtained through a teacher, can change a student's property e.g., color invariance to the same extent?%can distill knowledge such as color invariance into a student to the same extent? 

\textbf{Experimental setup:} We use ResNet18 as the student and train it for ImageNet classification using $KL$ method. However, for each input image $x$, instead of $\mathbf{z}_t$ (eq. 1, main paper) coming from an actual teacher, we generate the soft probabilities using $x$'s ground-truth label $\mathbf{y}$. We first add a random Gaussian noise with variance 0.2, and then perform the softmax operation with temperature 0.15 to convert it into a probability distribution. This probability vector then acts as the target for the student to match. We then evaluate the agreement score of this \emph{pseudo}-distilled student for color invariance (similar to Figure 4(b) in main paper).

\begin{wrapfigure}[11]{r}{0.4\textwidth}
	\centering
 	\vspace{-10pt}
	%\hspace*{5pt}
	\includegraphics[width=0.4\textwidth]{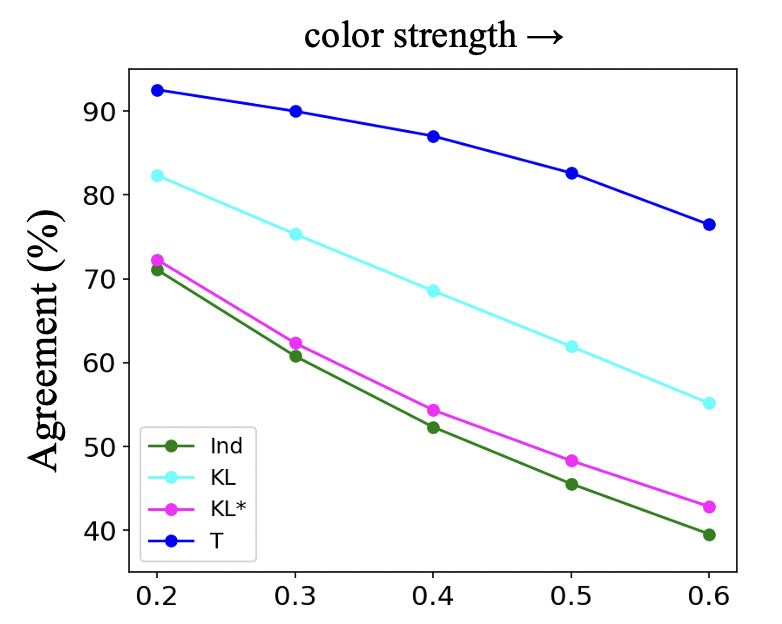}
	\vspace{-5pt}
\end{wrapfigure}
\textbf{Results:} We discuss three models, (i) the independent student ($Ind$): top-1 acc. = 70.04\%, (ii) student distilled using color-invariant ResNet50 as the teacher ($KL$): top-1 acc. = 71.10\%, and (iii) student distilled through the soft-labels without the teacher ($KL^{*}$): top-1 acc. = 70.49\%. 
%The top-1 accuracy of the models are: $Ind$: 70.04\%; $KL$: 71.1\%; $KL^{*}$: 70.49\%. So, we see that using soft label, even if it is not from a teacher, does improve the performance compared to the independent student. 
In the figure on the right, we see that while using soft-labels does marginally increase the agreement score of the student, it does not match the scores obtained by the students distilled with the actual color-invariant teacher. This reinforces the observation we made in section 4.3, that an increase in color invariance \emph{is primarily} due to certain knowledge being inherited from the teacher.  

\subsection{Does invariance to random crops transfer during knowledge distillation?}\label{sec:crop_inv}
This section extends the study done in Sec. 4 of the main paper, but for another popular data augmentation technique: randomly resized crops. 

\textbf{Experimental setup (crop invariance):} While training the teacher, we randomly crop the images as part of data augmentation (in addition to horizontal flips), with crop size between 8\% to 100\% of the image size. So, for example, the teacher can get to see a random 20\% region of an image in one iteration, and a random 80\% region of the same image in a different iteration. While training the students (independent or distilled), apart from horizontal flips, we only use center crop and \emph{do not} show random crops of an image.

\begin{wrapfigure}[13]{r}{0.26\textwidth}
	\centering
 	\vspace{-13pt}
	%\hspace*{5pt}
	\includegraphics[width=0.26\textwidth]{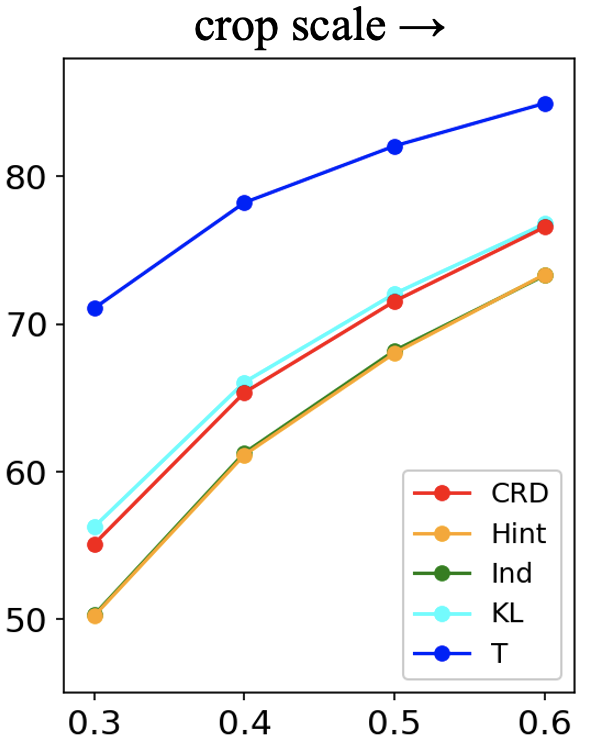}
	\vspace{-5pt}
\end{wrapfigure}
\textbf{Results (crop invariance):} During evaluation, we start with a test image $X$ from the 50k val set. We then set a crop scale, e.g. 0.2, and generate two random crops $X_1$ and $X_2$ so that both cover a random 20\% area of the original image $X$. Higher the crop scale, more image content will be common between the two crops. Then, we measure how frequently a model assigns the same class to $X_1$ and $X_2$. Fig. 4(d) (main paper) shows the agreement scores for increasing crop scales, where we again observe that the students distilled through $KL$ and $CRD$ become more invariant to this operation. Student distilled through $Hint$, however, does not increase its invariance to random crops, just as it did not increase its invariance to color jittering to the same extent as other methods in Fig. 4(b) (main paper).

\subsection{Does shift invariance transfer during knowledge distillation?}\label{sec:shift_inv}
Section 4 (main paper) and \ref{sec:crop_inv} (appendix) discussed whether invariance to certain data transformations can transfer from a teacher to the student during knowledge distillation. Fig.~\ref{fig:aug_vis} visualizes the effect of those transformations. Note that when we generate two random crops ($X_1, X_2$) of an image ($X$) with a fixed scale (e.g. 0.4), the aspect ratio of the two crops can still be kept different, which is what we do in Fig.~\ref{fig:aug_vis} (middle) and in the results shown in the previous section. If the aspect ratio is kept the same between $X_1$ and $X_2$, then one can study a more common property of neural networks: \emph{shift invariance} i.e. whether the network's predictions remain same if we shift  an image by certain pixels (either left/right/top/bottom). We study if this knowledge can be transferred from a teacher to the student during the distillation process.

\textbf{Experimental setup:} For the teacher, we choose a model which has been explicitly made to be shift-invariant. A recent work showed that a model's robustness to input shifts is related with the aliasing phenomenon, which refers to signal distorted with a small downsampling rate. To alleviate this issue and make CNNs shift invariant,~\cite{zhang2019shiftinvar} inserts low-pass filters into CNNs before downsampling. So, we use an anti-aliased ResNet50\# as the teacher (\# represents anti-aliased, same for the below). The student is the standard ResNet18 (without being anti-aliased). The distillation ResNet50\# $\rightarrow$ ResNet18 is done on the standard ImageNet dataset. The shift invariance of a model is evaluated across the 50k validation images in ImageNet. We start with a test image $X$ resized into 256x256 resolution. Then, we define the maximum shift we want in the resulting two images. If, for example, that value is 32, then we do a center crop of 256x256 followed by two random 224x224 crops to generate $X_1$ and $X_2$, keeping the aspect ratio same for both. If, instead, we desire a maximum shift of only 8 between $X_1$ and $X_2$, we would do a center crop of 232x232, followed by by two random 224x224 crops. Then, we compute how frequently a model gives the same prediction for $X_1$ and $X_2$, which is called the agreement score (same as section 4.3).

\begin{wrapfigure}[12]{r}{0.4\textwidth}
	\centering
 	\vspace{-13pt}
	%\hspace*{5pt}
	\includegraphics[width=0.4\textwidth]{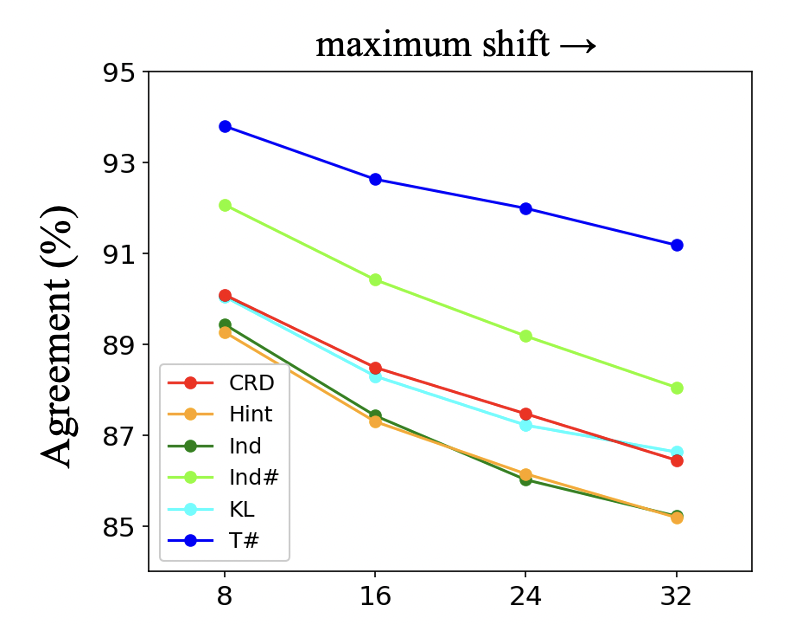}
	\vspace{-5pt}
\end{wrapfigure}
\textbf{Results:} In the figure on the right, we see the agreement scores of different models, and see that the agreement scores of the ResNet18 students distilled using KL and CRD increase relative to the independent ResNet18. Note that one can convert ResNet18 (the student) into its anti-aliased version as well by inserting low-pass filters~\cite{zhang2019shiftinvar}. The agreement score achieved by this student can be thought of as the upper-limit for a ResNet18 model, which we show by light green colored plot (denoted as Ind\#). Given the results of section 4.3 (crop-invariance), this result is expected since invariance to image shifts (aspect ratio constant) is a subset of invariance to random crops (aspect ratio could be different). Again, we observe that $Hint$ has difficulty in transferring this property.

\subsection{Does shape/texture bias get distilled?}\label{sec:shape_bias}
\vspace{-2pt}

% \begin{figure*}[t]
%     \centering
%     \includegraphics[width=1\textwidth]{figs/texture_crop.pdf}
%     \vspace{-5pt}
%     \caption{\textbf{Left} Shape biases of the ResNet18 models distilled from ResNet50 (* indicates the teacher is trained on Stylized-ImageNet) and ViT. Improvement can be seen when the teacher is ResNet50 (same family), something that is not observed with a transformer. \textbf{Right} Agreement scores of the models between $X_1$ and $X_2$ created using crop-augmentation. The two plots vary two different controls (crop size and crop ratio). Note the increase in agreement scores by $KL$ and $CRD$ compared to the independent student.}
%     \label{fig:shape_crop}
% \end{figure*}

The previous section dealt with knowledge about images from unseen domains, and the section before that discussed if certain invariances can be transferred. This section brings together those ideas to study an important property: shape/texture bias of neural networks. Prior work has shown that convolutional  networks tend to overly rely on texture cues when categorizing images \cite{cnn-texture}. 
%This has led to efforts being made to train models so that they have more shape bias, which is what humans have been found to use \cite{geirhos2021partial, cnn-texture, vit}. 
Here we study the following: If the teacher is shape biased, and the default (independent) student more texture biased, does distillation increase the shape bias of the distilled student? 

\textbf{Experimental setup:} We use the toolbox in \cite{geirhos2021partial} to compute the shape vs.~texture biases of a model. Shape bias is computed by using images with conflicting content and style information: e.g., an image with a shape (content) of a \emph{cat} but texture (style) of an \emph{elephant}. So, this particular image could have two correct decisions, a \emph{cat} or an \emph{elephant}. 
Using such images, the task is to see what fraction of correct decisions are based on shape vs.~texture information. 
%For teacher, we first chose ViT, since it was shown to have a high shape bias (0.615) without any explicit training \cite{geirhos2021partial}. 
%This follows a general trend that we have noticed in this work, where distilling knowledge from a transformer into a CNN architecture turns out to be difficult.
%The shape bias of the distilled students are shown in \ut{Fig. X}, where we see that there is no improvement across the different methods.
For the teacher, we choose a ResNet50$^{*}$ trained on Stylized-ImageNet \cite{cnn-texture}, where the image labels are kept the same, but the style is borrowed from arbitrary paintings. This way, the teacher has to focus more on shape information and consequently has a high shape bias of $\sim$0.81. We choose ResNet18 as the student, as it has a lower shape bias of $\sim$0.21. We then perform ResNet50$^{*}$ $\rightarrow$ ResNet18 distillation on the standard ImageNet dataset; i.e., the student is trained without any stylized images, while the teacher is, and we evaluate whether the student inherits the shape bias of the teacher. We also conduct an experiment with a transformer teacher and CNN student: ViT $\rightarrow$ ResNet18.  Since ViT have been shown to be inherently more shape-biased, we do not train the ViT teacher on Stylized-ImageNet, and instead train both it and the student on standard ImageNet.

%of two configurations in the figure on the right 
\textbf{Results:} For each distillation method, we show two results: one with lower weight on the distillation loss ($\downarrow$) and one with higher ($\uparrow$). From (a) in the right figure, we see that both $KL$ and $CRD$ improve the distilled student's shape bias, with a further jump obtained when using a higher weight, especially through $KL$. Sec. 4 (main paper) already showed that the student can indirectly inherit color invariance properties of the teacher. But, it is still interesting to see that, with proper hyperparameters, the inherited knowledge includes more subtle properties, like \emph{texture invariance} as well.

%\vspace{-50pt}
\begin{wrapfigure}[8]{r}{0.49\textwidth}
	\centering
 	\vspace{-10pt}
	%\hspace*{5pt}
	\includegraphics[width=0.49\textwidth]{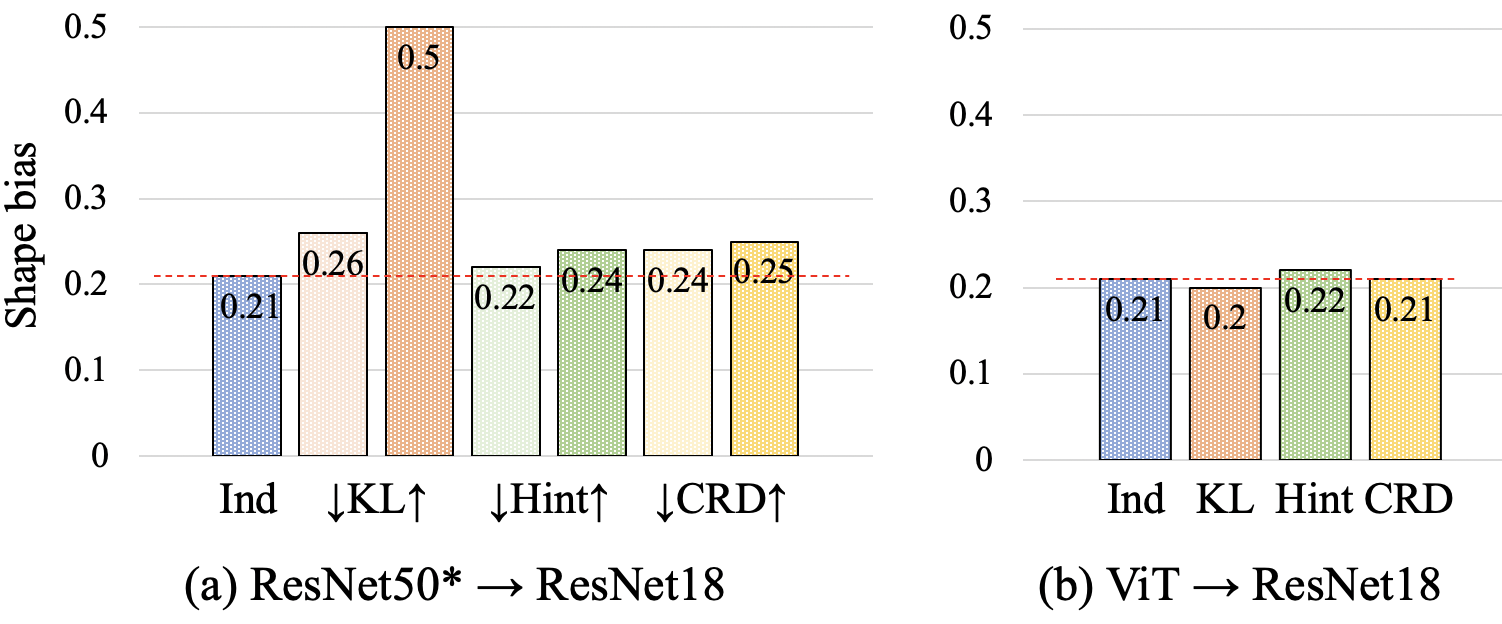}
	\vspace{-5pt}
\end{wrapfigure}
For ViT (shape bias = 0.615) $\rightarrow$ ResNet18, the shape bias of the distilled students do not change much (b). This follows a general trend where distilling knowledge from a transformer into a CNN turns out to be difficult. The implicit biases introduced due to architectural differences between the teacher and student, seem too big to be overcome by current distillation methods.

\subsection{Distillation makes internal representations to become similar}\label{sec:why}
\vspace{-2pt}

We hypothesize the following: when mimicking the teacher at a particular layer, the student's intermediate representations before that layer become similar as well.  That is, rather than predicting the activations in the target layer (e.g., output layer) in a very different way (e.g., the student classifying an image based on color features while the teacher classifies it based on shape), the student learns to behave more like the teacher throughout its network.  However, the degree to which this happens depends both on which layer the student mimics, and how similar the student's architecture is to that of the teacher. To study these aspects, we use centered kernel alignment (CKA) \cite{cka}, a popular method for measuring the similarity of two neural networks. Given two representations, $X \in \mathbb{R}^{n \times p_1}$ and $Y \in \mathbb{R}^{n \times p_2}$ of the same $n$ inputs, CKA(X, Y) $\in$ [0, 1] indicates how similar (close to 1) or dissimilar (close to 0) the two are. 

\textbf{Experimental setup:} We consider three settings: (i) ResNet50 $\rightarrow$ ResNet18 using $KD$; (ii) ResNet50 $\rightarrow$ ResNet18 using $Hint$ (distillation after the default second convolutional stage); and (iii) Swin-tiny $\rightarrow$ ResNet18 using $KD$. For each setting, we consider representations from (roughly) corresponding locations in the network (e.g., after the last layer in each convolutional stage). Seven corresponding locations are chosen from the teacher and student (for ResNets, the same layers used in the $Hint$ ablation study, Fig. 4d). We take 100 random images from the ImageNet validation set and compute their representations from those layers to construct a 7 x 7 similarity matrix. We compare the teacher to both the independent and distilled student to get two similarity matrices.

\textbf{Results:} Figure~\ref{fig:cka} shows the similarities between the teacher and the independent/distilled students. First, we see that the scores are higher between the corresponding feature representations (along the diagonal entries) of the distilled student and teacher networks for ResNet50 $\rightarrow$ ResNet18, with $KD$ resulting in a more significant gain than $Hint$. Second, we see very similar and low overall scores (except for the target F7 layer) for the independent and distilled students for Swin-tiny $\rightarrow$ ResNet18.  These support our hypothesis that the student learns similar intermediate representations as the teacher before the target layer, if the student and teacher's architectures are of the same family (e.g., both are ResNets).   Moreover, mimicking the output class probabilities ($KD$) leads to the student learning more similar representations as those of the teacher than mimicking an earlier layer ($Hint$). Finally,
when the architectures are very different (Swin-tiny and ResNet18), the intermediate representations do not become similar (despite a performance gain of the distilled student) because their inductive biases lead to different ways of learning the task.  Overall, our analysis shows that there is a correlation between the degree to which a student inherits the teacher's general properties and learned representation similarities.

\subsection{Results on additional datasets}
Beyond the results on ImageNet and MNIST, we study if the phenomena of implicit knowledge transfer exists for some other datasets as well. First, we conduct the adversarial vulnerability experiment on CIFAR-100~\cite{cifar100} (section 4.2 in the main paper). We report the results on three different teacher-student settings: (i) Wide ResNet 40-2 (WRN-40-2) -> ShuffleNetV1, (ii) VGG13 -> VGG8, (iii) ResNet50 -> MobileNetV2. Both the teacher and the students are trained on the training split of CIFAR-100, and tested on 5000 random images from the test split. The results shown below in Table~\ref{tab:cifar} depict the fooling rates (in \%) of different kinds of students when using adversarial images crafted for the teacher. We see that the fooling rates increase for distilled students, following a similar pattern as in Sec. 4.2 - Fig. 3. So, adversarial vulnerability of the teacher does get distilled into the students trained with different distillation objectives.

\begin{table}
\centering
\scriptsize
\begin{tabular}{l|ccc}

\toprule
%\cline{2-6}
                        & \textbf{WRN 40-2 $\rightarrow$ ShuffleNetV1} & \textbf{VGG-13 $\rightarrow$ VGG-8} & \textbf{ResNet50 $\rightarrow$ MobileNetV2}  \\ 
\midrule
\multicolumn{1}{l|}{Ind} &   31.23 & 42.42	& 36.57             \\ %\hline
\multicolumn{1}{l|}{KL} &   48.62 &	51.87 &	43.32                   \\ %\hline
\multicolumn{1}{l|}{Hint} &  62.63	& 49.79  &	43.91                    \\ %\hline
\multicolumn{1}{l|}{CRD} &  49.41 &	54.68	& 46.07                    \\ %\hline
\end{tabular}
\vspace{3pt}
\caption{Aversarial vulnerability results on CIFAR-100 (analogous to Section~\ref{sec:adv} in the main paper.}
\label{tab:cifar}
\end{table}

Next, similar to the MNIST domain adaptation experiment in Section~\ref{sec:good_mnist}, we conduct the experiment on a more real-world domain. We consider two datasets: VLCS and PACS ~\cite{li-domain}. VLCS consists of images from four domains - VOC2007, LabelMe, Caltech-101, and SUN - where in each domain there are images belonging to five categories. PACS consists of images from four domains as well - sketch, photo, cartoon, art painting. Each domain consists of the same seven object categories. The teacher is trained on all the four domains, but the students (independent and distilled) are trained on three domains (images from one domain are never shown). The unseen domains are Caltech 101 and Photo when working with VLCS and PACS datasets respectively.

The goal is to see if mimicking the teacher on three domains also helps the student inherit teacher’s information on the fourth (hidden) domain; i.e., do we see an improvement in the accuracy on that hidden domain for the distilled students, compared to an independent student. The results, depicting the classification accuracy of different models, are shown below in Tables~\ref{tab:vlcs} and \ref{tab:pacs}. The distilled students’ performance improves on the unseen domains in both the cases, simply by having access to teacher’s responses on the other three domains. This is particularly pronounced when distillation is done using KL. For example, for VLCS, the performance on the unseen domain (Caltech 101) improves from 54.31 by the independent student to 71.83 by the KL distilled student.

\begin{table}
\centering
\scriptsize
\begin{tabular}{l|cccc}

\toprule
%\cline{2-6}
                        & \textbf{Caltech 101 (unseen)} & \textbf{LabelMe (seen)} & \textbf{SUN09 (seen)} & \textbf{VOC 2007 (seen)}  \\ 
\midrule
\multicolumn{1}{l|}{Ind} & 54.31 &	61.07 &	60.38 &	51.94               \\ %\hline
\multicolumn{1}{l|}{KL} & 71.83 &	60.73 &	61.52 &	53.12                     \\ 
\multicolumn{1}{l|}{Hint} &  67.95 & 60.84 & 60.34 & 52.76                    \\ 
\multicolumn{1}{l|}{CRD} &   61.66 & 63.57 & 55.31 & 53.12                  \\ %\hline
\end{tabular}
\vspace{3pt}
\caption{Domain adaptation results on VLCS results (analogous to Fig.~\ref{fig:mnist}). Teacher's knowledge about the unseen domain (Caltech 101) gets transferred, to some extent, into the distilled student.}
\label{tab:vlcs}
\end{table}

\begin{table}[h]
\centering
\scriptsize
\begin{tabular}{l|cccc}

\toprule
%\cline{2-6}
                        & \textbf{Photo (unseen)} & \textbf{Sketch (seen)} & \textbf{Cartoon (seen)} & \textbf{Art (seen)}  \\ 
\midrule
\multicolumn{1}{l|}{Ind} &  40.91 &	70.17 &	66.47 &	45.56              \\ %\hline
\multicolumn{1}{l|}{KL} & 49.93 &	72.82 &	69.69 &	46.64                     \\ 
\multicolumn{1}{l|}{Hint} & 48.34 & 71.13 &	68.08 &	45.89                     \\ 
\multicolumn{1}{l|}{CRD} &  44.49 &	71.52 &	67.14 &	47.86                   \\ %\hline
\end{tabular}
\vspace{3pt}
\caption{Domain adaptation results on PACS dataset (analogous to Table~\ref{tab:vlcs}).}
\label{tab:pacs}
\end{table}

\section{Supporting quantitative results}\label{sec:quant_res}
Finally, we report the performance of different models on ImageNet 50k validation set. Table~\ref{tab:acc} lists the top-1 accuracies of different models used in the main paper. Overall, we have tried to use the hyper-parameters which improve the distilled student's performance compared to the independent student. In every case, we use a single teacher to perform distillation into two students trained with different random seeds i.e. Teacher $\rightarrow$ Student$_{1}$ and Teacher $\rightarrow$ Student$_{2}$, for each method. We then report the results shown in the main paper with their respective error bars, in Tables~\ref{tab:adv_std}-\ref{tab:mnist_std}.

%Wherever possible, we have used teacher as the pre-trained model from PyTorch~\cite{pytorch}. Note that in certain cases, e.g. ResNet50 (sty) $\rightarrow$ ResNet18

\begin{table}[h!]
\footnotesize
\scriptsize
\centering
\begin{tabular}{l|ccccc}
\toprule
%\small
%\cline{2-6}
                        & \textbf{Teacher} & \textbf{Ind} & \textbf{KL} & \textbf{Hint} & \textbf{CRD} \\ \hline
\multicolumn{1}{l|}{ResNet50 $\rightarrow$ ResNet18} &   76.13   &    70.04$\pm$0.01  & 70.98$\pm$0.01   &   70.56$\pm$0.16   &    70.73$\pm$0.02  \\ %\hline

\multicolumn{1}{l|}{VGG19 $\rightarrow$ VGG11} &   72.37      &   68.88$\pm$0.01          &  69.74$\pm$0.10  &  69.38$\pm$0.15    &   69.74$\pm$0.07  \\ %\hline

\multicolumn{1}{l|}{VGG19 $\rightarrow$ ResNet18} &  72.37    &    70.04$\pm$0.01   &  70.62$\pm$0.02  &   70.21$\pm$0.30   &  70.42$\pm$0.07   \\ %\hline

\multicolumn{1}{l|}{ViT $\rightarrow$ ResNet18} &   75.91      &    70.04$\pm$0.01         & 70.39$\pm$0.02   &   70.59$\pm$0.07   &  70.58$\pm$0.03   \\ %\hline

\multicolumn{1}{l|}{Swin-Base $\rightarrow$ Swin-Tiny} &  83.50       &  81.13$\pm$0.08           &  81.23$\pm$0.04   &  81.33$\pm$0.11    &  81.27 $\pm$0.21   \\ %\hline

\multicolumn{1}{l|}{ResNet50 (sty) $\rightarrow$ ResNet18  $\uparrow$} &  60.18        &      70.04$\pm$0.01       & 61.45$\pm$0.07     &  68.82$\pm$0.12    &  69.56$\pm$0.07   \\ %\hline

\multicolumn{1}{l|}{ResNet50 (sty) $\rightarrow$ ResNet18  $\downarrow$} &  60.18        &      70.04$\pm$0.01       & 70.65$\pm$0.03     &  70.45$\pm$0.07    &  69.96$\pm$0.05   \\ %\hline

\multicolumn{1}{l|}{ResNet50 (col) $\rightarrow$ ResNet18} &  75.32       &     70.04$\pm$0.01        &    71.01$\pm$0.06 &   70.41$\pm$0.20     &     70.97$\pm$0.19  \\ %\hline

\multicolumn{1}{l|}{ResNet50 $\rightarrow$ ResNet18 (w/o crop)} &   76.13      &     64.84$\pm$0.02        &  68.75$\pm$0.01  &  64.81$\pm$0.14    &  67.41$\pm$0.07  \\ %\hline

\end{tabular}
\vspace{3pt}
\caption{Top-1 accuracy (in \%) of different models on 50k ImageNet validation images.}
\label{tab:acc}
\end{table}

%\section{Results with error bars}

\begin{table}[h!]
\centering
\scriptsize
\begin{tabular}{l|ccccc}
\toprule
%\cline{2-6}
                        & \textbf{Teacher} & \textbf{Ind} & \textbf{KL} & \textbf{Hint} & \textbf{CRD} \\ \hline
\multicolumn{1}{l|}{ResNet50 $\rightarrow$ ResNet18} &    84.82     &     44.16$\pm$0.19        &  51.98$\pm$2.44   &  48.34$\pm$0.34    &  50.46$\pm$0.29   \\ %\hline
\multicolumn{1}{l|}{VGG19 $\rightarrow$ VGG11} &   87.22      &     62.29$\pm$0.36        &  69.74$\pm$0.67   &  79.78$\pm$0.08    &  70.51$\pm$0.78   \\ %\hline
\multicolumn{1}{l|}{VGG19 $\rightarrow$ VGG11 (R18)} &    87.22     &   69.02$\pm$0.48   &  70.54$\pm$0.90  &   70.68$\pm$0.62 &  70.59$\pm$0.62   \\ %\hline
\multicolumn{1}{l|}{ViT $\rightarrow$ ResNet18} &    85.84     &    21.93$\pm$0.24  &  21.57$\pm$0.49    &  23.34$\pm$0.14     &   23.47$\pm$0.31    \\ %\hline
\multicolumn{1}{l|}{VGG19 $\rightarrow$ ResNet18} &     87.22    &     36.19$\pm$0.01        &  43.02$\pm$0.06  &   47.68$\pm$0.47     &  48.99$\pm$0.05   \\ %\hline

\end{tabular}
\vspace{3pt}
\caption{Adversarial fooling rates (in \%), corresponding to Figure 3 in the main paper. 
}
\label{tab:adv_std}
\end{table}

\begin{table}[h!]
\centering
\scriptsize
\begin{tabular}{l|cc}
\toprule
%\cline{2-6}
                        & \textbf{ResNet50 (col)} & \textbf{ResNet50}  \\ \hline
\multicolumn{1}{l|}{Ind} &     71.27$\pm$0.21    &      71.27$\pm$0.21          \\ %\hline
\multicolumn{1}{l|}{KL} &   82.10$\pm$0.07     &    74.02$\pm$0.23            \\ %\hline
\multicolumn{1}{l|}{Hint} &    72.22$\pm$0.14     &       72.42$\pm$0.39        \\ %\hline
\multicolumn{1}{l|}{CRD} &   79.44$\pm$0.20      &     71.27$\pm$0.25          \\ %\hline
\end{tabular}
\vspace{3pt}
\caption{Table corresponding to Figure 4(a) in main paper. Knowledge transfer about color information from two teachers: color invariant ResNet50 (T) and default ResNet50 (T$^{*}$).}

\scriptsize
\centering
\begin{tabular}{l|cccc}
\toprule
%\cline{2-6}
                        & \textbf{0.3} & \textbf{0.4} & \textbf{0.5} & \textbf{0.6}  \\ \hline
\multicolumn{1}{l|}{Ind} &    60.77$\pm$0.10     &    52.32$\pm$0.21    &   45.55$\pm$0.20   &     39.56$\pm$0.31        \\ %\hline
\multicolumn{1}{l|}{KL} &    75.32$\pm$0.17     &  68.56$\pm$0.36   &   61.93$\pm$0.33    &    55.15$\pm$0.37            \\ %\hline
\multicolumn{1}{l|}{Hint} &     61.96$\pm$0.00    &   53.34$\pm$0.41    &   47.72$\pm$0.51    &    42.00$\pm$0.53           \\ %\hline
\multicolumn{1}{l|}{CRD} &    71.83$\pm$0.48     &   64.31$\pm$0.14     &   57.26$\pm$0.47    &    49.90$\pm$0.48           \\ %\hline
\end{tabular}
\vspace{3pt}
\caption{Table corresponding to Figure 4(b) in main paper. Illustration of knowledge transfer in, ResNet50 $\rightarrow$ ResNet18, if the two images have increasingly different color properties.}

\scriptsize
\centering
\begin{tabular}{l|cccc}
\toprule
%\cline{2-6}
                        & \textbf{0.3} & \textbf{0.4} & \textbf{0.5} & \textbf{0.6}  \\ \hline
\multicolumn{1}{l|}{Ind} &    60.77$\pm$0.10     &    52.32$\pm$0.21    &    45.55$\pm$0.20   &      39.56$\pm$0.31       \\ %\hline
\multicolumn{1}{l|}{KL} &    62.49$\pm$0.09     &   54.18$\pm$0.15  &    47.68$\pm$0.50   &     41.92$\pm$0.48           \\ %\hline
\multicolumn{1}{l|}{Hint} &    61.68$\pm$0.04     &    53.63$\pm$0.19   &    47.37$\pm$0.32   &      41.76$\pm$0.50         \\ %\hline
\multicolumn{1}{l|}{CRD} &   60.85$\pm$0.65      &    52.80$\pm$0.76    &   46.91$\pm$1.29    &     42.00$\pm$1.31          \\ %\hline
\end{tabular}
\vspace{3pt}
\caption{Table corresponding to Figure 4(c) in main paper. Illustration of knowledge transfer in, Swin-Tiny $\rightarrow$ ResNet18, if the two images have increasingly different color properties.}

\scriptsize
\centering
\begin{tabular}{l|cccc}
\toprule
%\cline{2-6}
                        & \textbf{0.3} & \textbf{0.4} & \textbf{0.5} & \textbf{0.6}  \\ \hline
\multicolumn{1}{l|}{Ind} &    50.30$\pm$0.21     &   61.27$\pm$0.10     &    68.20$\pm$0.50   &     73.30$\pm$0.33        \\ %\hline
\multicolumn{1}{l|}{KL} &    56.26$\pm$0.00     &   66.06$\pm$0.02  &    72.06$\pm$0.19   &     76.79$\pm$0.04           \\ %\hline
\multicolumn{1}{l|}{Hint} &    50.23$\pm$0.27     &    61.14$\pm$0.11   &    68.04$\pm$0.14   &     73.36$\pm$0.08          \\ %\hline
\multicolumn{1}{l|}{CRD} &    55.10$\pm$0.12     &   65.36$\pm$0.43     &   71.55$\pm$0.26    &     76.57$\pm$0.43          \\ %\hline
\end{tabular}
\vspace{3pt}
\caption{Table corresponding to Figure 4(d) in main paper. Illustration of knowledge transfer in, ResNet50 $\rightarrow$ ResNet18, if the two images are random crops of increasing scales. 
%Lower the crop scale (size), more different the two random crops will be.
}

\end{table}

\begin{table}[h]
\centering
\scriptsize
\begin{tabular}{l|llllll}
\toprule
%\hline
     & \multicolumn{6}{c}{VGG19 $\rightarrow$ ResNet18}                                                                                                                                                                                               \\ \cline{2-7}
     & \multicolumn{1}{c|}{\scriptsize \textbf{sketch}} & \multicolumn{1}{c|}{\scriptsize \textbf{stylized}} & \multicolumn{1}{c|}{\scriptsize \textbf{silhouette}} & \multicolumn{1}{c|}{\scriptsize \textbf{edge}} & \multicolumn{1}{c|}{\scriptsize \textbf{cue conflict}} & \scriptsize \textbf{ImageNet val}  \\ \hline
Ind  & \multicolumn{1}{c|}{33.62$\pm$0.12}       & \multicolumn{1}{c|}{21.68$\pm$0.19}         & \multicolumn{1}{c|}{12.81$\pm$0.94}           & \multicolumn{1}{c|}{26.25$\pm$1.25}     &      \multicolumn{1}{c|}{22.81$\pm$0.00}      &      75.60$\pm$0.01 \\ %\hline
KL  & \multicolumn{1}{c|}{37.56$\pm$0.31}       & \multicolumn{1}{c|}{28.81$\pm$0.44}         & \multicolumn{1}{c|}{31.25$\pm$5.00}           & \multicolumn{1}{c|}{31.25$\pm$5.00}     &      \multicolumn{1}{c|}{29.30$\pm$2.03}       &       77.21$\pm$0.06  \\ %\hline
Hint  & \multicolumn{1}{c|}{37.18$\pm$1.19}       & \multicolumn{1}{c|}{27.19$\pm$0.19}         & \multicolumn{1}{c|}{10.00$\pm$3.75}           & \multicolumn{1}{c|}{29.69$\pm$2.19}     &     \multicolumn{1}{c|}{27.73$\pm$1.25}      &       76.49$\pm$0.09 \\ %\hline
CRD  & \multicolumn{1}{c|}{40.50$\pm$0.37}       & \multicolumn{1}{c|}{30.75$\pm$0.50}         & \multicolumn{1}{c|}{37.81$\pm$2.19}           & \multicolumn{1}{c|}{35.00$\pm$1.25}     &     \multicolumn{1}{c|}{30.93$\pm$0.08}       &       78.36$\pm$0.06 \\ 
\end{tabular}
\vspace{7pt}

\begin{tabular}{l|llllll}
%\hline
\toprule
     & \multicolumn{6}{c}{Swin-Base $\rightarrow$ Swin-Tiny}                                                                                                                                                                                               \\ \cline{2-7}
     & \multicolumn{1}{c|}{\scriptsize \textbf{sketch}} & \multicolumn{1}{c|}{\scriptsize \textbf{stylized}} & \multicolumn{1}{c|}{\scriptsize \textbf{silhouette}} & \multicolumn{1}{c|}{\scriptsize \textbf{edge}} &  \multicolumn{1}{c|}{\scriptsize \textbf{cue conflict}} &  \scriptsize \textbf{ImageNet val}  \\ \hline
Ind  & \multicolumn{1}{c|}{51.37$\pm$0.37}       & \multicolumn{1}{c|}{33.75$\pm$0.37}         & \multicolumn{1}{c|}{22.50$\pm$1.25}           & \multicolumn{1}{c|}{50.00$\pm$0.00}     &    \multicolumn{1}{c|}{37.26$\pm$0.47}    &   88.79$\pm$0.07  \\ %\hline
KL  & \multicolumn{1}{c|}{56.93$\pm$1.06}       & \multicolumn{1}{c|}{38.43$\pm$0.68}         & \multicolumn{1}{c|}{27.50$\pm$2.50}           & \multicolumn{1}{c|}{57.81$\pm$1.56}     &    \multicolumn{1}{c|}{42.61$\pm$0.04}    &  89.39$\pm$0.05  \\ %\hline
Hint  & \multicolumn{1}{c|}{52.56$\pm$1.18}       & \multicolumn{1}{c|}{35.18$\pm$0.19}         & \multicolumn{1}{c|}{26.87$\pm$1.87}           & \multicolumn{1}{c|}{54.37$\pm$2.50}     &     \multicolumn{1}{c|}{38.51$\pm$0.62}   &  89.03$\pm$0.17 \\ %\hline
CRD  & \multicolumn{1}{c|}{54.18$\pm$0.94}       & \multicolumn{1}{c|}{34.75$\pm$1.50}         & \multicolumn{1}{c|}{26.25$\pm$0.00}           & \multicolumn{1}{c|}{50.62$\pm$0.00}     &    \multicolumn{1}{c|}{39.22$\pm$0.47}     &  88.97$\pm$0.01  \\ 
%. 
\end{tabular}

\vspace{5pt}
\caption{Consensus scores between teacher and the student, corresponding to Figure 5 in the paper. ImageNet val denotes the 50k images in the validation set of the seen domain (ImageNet).}
\label{tab:ood_std_ref}
\end{table}

\begin{table}[h]
\centering
\begin{tabular}{l|cc|c}
\toprule

%\hline
     & \multicolumn{2}{c|}{ResNet50 (sty) $\rightarrow$ ResNet18}           & \multicolumn{1}{c}{\multirow{2}{*}{ViT $\rightarrow$ ResNet18}} \\ \cline{2-3}
     & \multicolumn{1}{c|}{Lower} & \multicolumn{1}{c|}{Higher} & \multicolumn{1}{c}{}                                \\ \hline
Ind  & \multicolumn{1}{l|}{0.21$\pm$0.01}      &            0.21$\pm$0.01                 &             0.21$\pm$0.01                                         \\ %\hline
KL   & \multicolumn{1}{l|}{0.26$\pm$0.01}      &             0.50$\pm$0.00                &              0.20$\pm$0.01                                        \\ %\hline
Hint & \multicolumn{1}{l|}{0.22$\pm$0.01}      &           0.24$\pm$0.02                  &               0.22$\pm$0.00                                       \\ %\hline
CRD  & \multicolumn{1}{l|}{0.24$\pm$0.00}      &           0.25$\pm$0.01                  &               0.21$\pm$0.00                                       \\ %\hline
\end{tabular}
\caption{Shape bias scores of students, corresponding to the figure in Section 4.5 in the main paper.}
\end{table}

\begin{table}[h]
\centering
\begin{tabular}{l|ccc}
\toprule
     & MNIST-orig & MNIST-Color & MNIST-M  \\ \hline
Ind  &      99.08$\pm$0.07      &    72.86$\pm$2.32         &     56.09$\pm$1.14    \\
KL   &      98.90$\pm$0.01      &    91.76$\pm$1.00         & 67.92$\pm$1.07        \\
Hint &       99.10$\pm$0.06     &        97.05$\pm$0.05     &     64.06$\pm$0.93    \\
CRD  &     99.00$\pm$0.10       &      83.98$\pm$0.88       &      60.36$\pm$0.23  
\end{tabular}
\vspace{3pt}
\caption{Top-1 accuracy of distilled models, corresponding to figure 6 in the main paper.}
\label{tab:mnist_std}
\end{table}

%\newpage

% {
% \bibliographystyle{iclr2023_conference}
% \bibliography{main}
% }

\begin{table}[t!]
\centering
\begin{tabular}{l|cc}
\toprule
       & $\mathcal{D}_s$ & $\mathcal{D}_t$ \\ \hline
Race 1 & 600  & 4000 \\
Race 2 & 50   & 4000 \\
Race 3 & 2000 & 0    \\
Race 4 & 200  & 4000 \\
Race 5 & 0    & 4000 \\
Race 6 & 200  & 4000 \\
Race 7 & 800  & 4000
\end{tabular}
\caption{Dataset composition of FairFace~\cite{fairface}. Different rows represent the number of training images used from each race.}
\label{tab:ff_dataset}
\end{table}

\end{document}